\pdfoutput=1

\documentclass[11pt]{article}

\usepackage[preprint]{acl}

\usepackage{times}
\usepackage{latexsym}

\usepackage[T1]{fontenc}

\usepackage[utf8]{inputenc}

\usepackage{microtype}

\usepackage{inconsolata}

\usepackage{graphicx}

\usepackage{amsmath,amsfonts,bm}

\def\eqref#1{equation~\ref{#1}}

\def\1{\bm{1}}

\DeclareMathAlphabet{\mathsfit}{\encodingdefault}{\sfdefault}{m}{sl}
\SetMathAlphabet{\mathsfit}{bold}{\encodingdefault}{\sfdefault}{bx}{n}

\usepackage{amsmath,amsfonts,bm}
\usepackage{booktabs} %
\usepackage{url}
\usepackage{graphicx}
\usepackage{array}
\usepackage{color}
\usepackage{makecell}
\usepackage{multirow}
\usepackage{xspace}
\usepackage{colortbl}
\usepackage{subcaption}
\usepackage{algpseudocode}
\usepackage[noend,ruled]{algorithm2e}
\usepackage{setspace}
\usepackage{varwidth}
\usepackage{enumitem}
\usepackage{mdframed}
\usepackage{centernot}
\usepackage{arydshln}

\usepackage{soul}
\usepackage{pifont}
\usepackage{graphics}

\usepackage{wrapfig}

\usepackage{cuted,tcolorbox,lipsum}

\newcommand{\modelname}{\textsc{FenCE}\xspace}

\newcommand{\circled}[1]{\textcircled{\raisebox{-0.9pt}{#1}}}

\newcommand{\start}[1]{\vspace{.3mm}\noindent{{\bf #1}.}}

\newcommand{\downv}{\vspace{-.1cm}}
\newcommand{\upv}{\vspace{-.1cm}}

\definecolor{amber}{rgb}{1.0, 0.75, 0.0}
\definecolor{applegreen}{rgb}{0.55, 0.71, 0.0}
\definecolor{treegreen}{rgb}{0.13, 0.54, 0.23}
\definecolor{LightCyan}{rgb}{0.88,1,1}
\definecolor{LightBlue}{RGB}{171, 227, 235}

\newcommand{\dlrow}{\rowcolor{gray!20}}
\newcommand{\blrow}{\rowcolor{LightBlue!30}}

\definecolor{green2}{HTML}{BFD8B6}
\definecolor{green3}{HTML}{E7F0E5}
\definecolor{greenarrow}{HTML}{1DB100}
\definecolor{red3}{HTML}{C82506}

\definecolor{gred}{RGB}{255,102,102}
\definecolor{gblue}{RGB}{51,102,255}
\definecolor{gyellow}{RGB}{244,180,0}
\definecolor{ggreen}{RGB}{15,157,88}
\definecolor{ggrey}{RGB}{115,115,115}
\definecolor{na}{gray}{0.9}

\definecolor{textRed}{RGB}{157,0,23}
\definecolor{textYellow}{RGB}{166,119,54}
\definecolor{textGreen}{RGB}{58,110,38}
\definecolor{textBlue}{RGB}{39,71,156}

\definecolor{LightYellow}{RGB}{255,250,208}
\definecolor{LightGreen}{RGB}{194,255,192}
\definecolor{LightBlue}{RGB}{187,236,251}
\definecolor{LightPurple}{RGB}{224,223,255}
\definecolor{LightGrey}{RGB}{225,225,225}

\definecolor{Grey}{RGB}{150,150,150}

\definecolor{OrangeRed}{rgb}{1.0, 0.27, 0.0}
\definecolor{midnightgreen}{rgb}{0.0, 0.29, 0.33}
\definecolor{darkgreen}{rgb}{0.0, 0.42, 0.24}

\definecolor{diagramRed}{RGB}{246,193,193}
\definecolor{diagramPurple}{RGB}{224,224,253}
\definecolor{diagramOrange}{RGB}{244,222,176}

\newcommand{\colorR}[1]{\textcolor{gred}{\textbf{#1}}}
\newcommand{\colorG}[1]{\textcolor{ggreen}{\textbf{#1}}}

\newcommand{\colorgrey}[1]{\textcolor{Grey}{#1}}

\usepackage{hyperref}
\usepackage{url}

\title{Improving Model Factuality with Fine-grained Critique-based Evaluator}

\author{
Yiqing Xie$^{1}$
\quad Wenxuan Zhou$^{2}\thanks{Equal contribution.}$
\quad Pradyot Prakash$^{2*}$
\quad {\bf Di Jin}$^{2}$
\quad {\bf Yuning Mao}$^{2}$ \\
\quad {\bf Quintin Fettes}$^{2}$
\quad {\bf Arya Talebzadeh}$^{2}$
\quad {\bf Sinong Wang}$^{2}$
\quad {\bf Han Fang}$^{2}$ \\
\quad {\bf Carolyn Rosé}$^{1}$
\quad {\bf Daniel Fried}$^{1,2}$
\quad {\bf Hejia Zhang}$^{2}$ \\
\\
$^{1}$Carnegie Mellon University 
\quad $^{2}$Meta GenAI 
}

\begin{document}
\maketitle
\begin{abstract}
Factuality evaluation aims to detect factual errors produced by language models (LMs) and hence guide the development of more factual models.
Towards this goal, we train a factuality evaluator, \modelname, that provides LM generators with claim-level factuality feedback.
In particular, we train \modelname to (1) generate textual critiques along with scores and (2) make claim-level judgment based on diverse source documents obtained by various tools, via data augmentation on a combination of \textit{public} judgment datasets.
We then present a framework that leverages \modelname to improve the factuality of LM generators by constructing training data.
Specifically, we generate a set of candidate responses, ask \modelname to revise and score each response without introducing lesser-known facts, and train the generator by preferring highly scored revised responses. 
Experiments show that our data augmentation methods improve the evaluator's accuracy by 2.9\% on LLM-AggreFact.
With \modelname, we improve Llama2-7B-chat/Llama3-8B-chat's factuality rate by 16.86\%/14.45\% on FActScore, outperforming state-of-the-art factuality finetuning methods by 8.83\%/6.96\%.
\end{abstract}

\begin{figure*}[t]
\centering
    \centering
    \includegraphics[width=\linewidth]{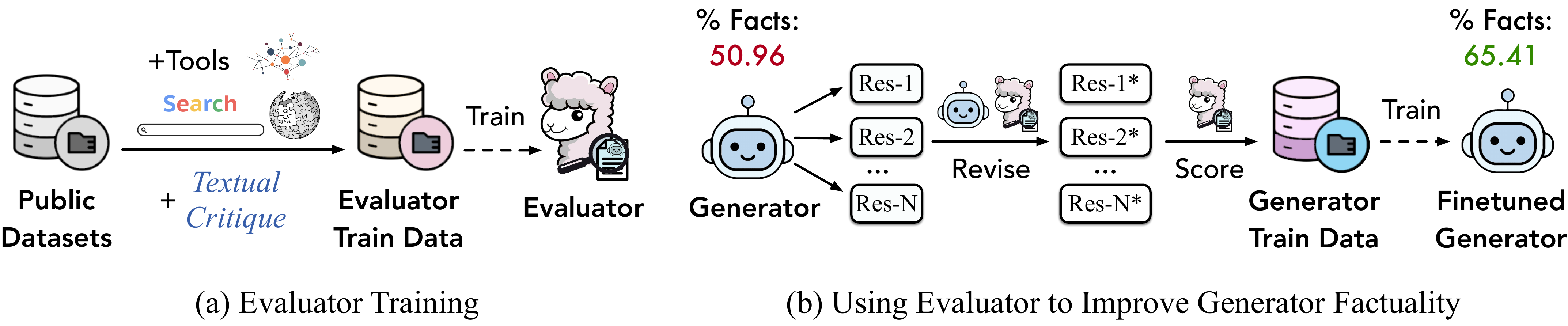}
    \upv
    \vspace{-0.2cm}
    \caption{\textit{(Left)} The framework to train an evaluator, \modelname, by augmenting public datasets with textual critiques and more diverse knowledge sources. We show the details in \autoref{fig:evaluator_train}.
    \textit{(Right)} The framework to improve model factuality with \modelname. We construct training data by leveraging \modelname to revise and score the generator's responses. Details of response revision are shown in \autoref{fig:revise}.}
    \label{fig:overview}
    \downv
\end{figure*}

\section{Introduction}
Hallucination is one of the persistent challenges for large language models (LLMs), where models generate plausible sounding but incorrect information, even if they are shown factual information during pretraining~\citep{hallucination-survey,empirical-factuality}. 
One hypothesis is that LLMs fail to distinguish the boundary between memorized facts and other plausible sounding information and do not learn to only output memorized facts, especially on their unfamiliar topics~\citep{finetuning-on-new,understand-finetuning,unfamiliar-finetuning}.
Although it is possible to reduce hallucination in inference with decoding strategies~\citep{ITI,DoLA} or post-editing~\citep{FAVA,EVER}, they introduce severe latency issues and hurts efficiency in real-time applications.

Alternatively, prior studies train the generator to output more factual responses, by preferring (1)  generation candidates with higher factuality scores \citep{FactTune}, which is limited by the generator's capabilities, or (2) responses with false information corrected~\citep{EVER}, 
which is prone to introducing lesser-known facts.
As shown in recent work~\citep{finetuning-on-new,understand-finetuning}, such preference training reinforces the model to generate information not well memorized during pretraining, and hence could even hurt factuality.
Furthermore, the methods either leverage proprietary models that have restricted terms of use, or prompt the generator to evaluate its own factuality, which suffers from self-bias and leads to inaccurate judgments~\citep{self-bias}.

Recent work trains evaluator models that could potentially be used to provide training signals for generators.
One category of work relies on proprietary models (e.g., GPT-4) to generate training data in various formats~\citep{prometheus,AutoJ}.
In contrast, \citet{FLAMe} leverage public datasets containing judgments of whether a claim is factual against certain source documents. 
However, such documents are generally sampled from very \textbf{restricted sources} such as news corpora or Wikipedia, while an evaluator could potentially benefit from knowledge obtained by a multiplicity of tools (e.g., search engines)~\citep{longform}.
Furthermore, the judgment label in most datasets is a single binary or numeric score, providing \textbf{limited feedback} to the generator model.

In this paper, we present \modelname, a \textbf{F}in\textbf{e}-grai\textbf{n}ed \textbf{C}ritic-based \textbf{E}valuator that aims to provide textual critiques for each model-generated claim based on diverse knowledge sources.
We start with a set of public datasets with human judgments on the factuality of model-generated claims. 
As shown in \autoref{fig:overview}(a), we augment the judgment labels with textual critiques, which provides more informative and explainable feedback to the generator.
In addition, we augment the source documents by invoking multiple tools, including a search engine, knowledge base, and knowledge graph, 
with the goal of training the evaluator to leverage more diverse knowledge sources.

We further demonstrate how to leverage \modelname to improve the generator's factuality with finetuning.
To construct training data (\autoref{fig:overview}(b)), we generate multiple responses for each prompt, use \modelname to judge and critique every claim in the responses, and replace false information with facts or remove it from the response, depending on whether the corresponding fact is within the generator's knowledge 
(i.e., whether the generator outputs \textit{``unknown''} when prompted about claim correctness). 
This mitigates introducing lesser-known facts into training data.
Finally, unlike existing work that use the generator itself as the evaluator, we use \modelname to score each original and revised response and construct preference data, which reduces self-bias and produces more accurate judgments.

Experiments show that \modelname outperforms large open-source models such as Mistral-123B and strong proprietary models such as Claude-3 on LLM-Aggrefact.
With \modelname, we improve Llama2-7B-chat's factuality by 16.86\% on FActScore~\citep{factscore} and 17.64\% on TruthfulQA \cite{truthfulqa}, outperforming existing factuality finetuning methods by 8.83\% and 3.99\%, respectively.
Analyses further show that after training with our recipe, the generator outputs less information for unfamiliar entities and more information for popular ones, suggesting that it learns to only generate information that is likely to be factual.

\start{Contributions}
(1) We train a fine-grained critique-based evaluator, \modelname, by augmenting public datasets with textual critiques and more diverse source documents.
(2) We propose a training recipe to improve the generator's factuality by leveraging \modelname to improve and score its responses.
(3) We conduct extensive experiments to validate both the judgment accuracy of \modelname and the factuality of the generator trained with \modelname, outperforming state-of-the-art factuality training methods by 8.83\% on FActScore and 3.99\% on TruthfulQA.

\section{Methodology}

\begin{figure*}[t]
\centering
    \centering
    \includegraphics[width=\linewidth]{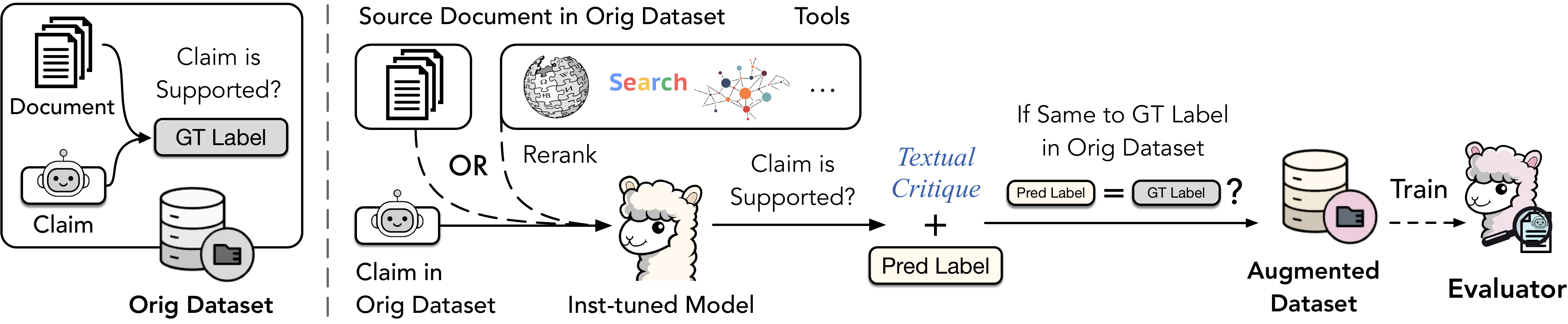}
    \upv
    \vspace{-0.2cm}
    \caption{Framework of evaluator training. (\textit{left}) Existing public datasets for evaluator training. Each example contains a claim, a source document, and the ground truth (GT) label of whether the claim is supported by the document. (\textit{Right}) We augment the datasets with textual critiques and more diverse source documents obtained by tools. We use zero-shot Llama-3-70B-chat as the instruction-tuned model in our experiments.}
    \label{fig:evaluator_train}
    \downv
\end{figure*}

\subsection{\modelname: Factuality Evaluator Training}
\label{sec:evaluator_training}

\start{Preliminary}
To train the evaluator to recognize hallucinations, previous work~\citep{FLAMe} incorporates a combination of datasets with human judgments on factuality. As shown in \autoref{fig:evaluator_train}, in most datasets, each example contains a claim, a source document, and the ground-truth judgement of whether the claim is (1) fully supported by, (2) contradicted with, or (3) contains information that cannot be verified by the document.

We formally define the problem as follows:
Given a claim $c \in \mathcal{C}$, a source document $d \in \mathcal{D}$, a factuality evaluator aims to learn a mapping $f: \mathcal{C} \times \mathcal{D} \rightarrow \mathcal{L}$, which maps each claim-document pair $(c,d)$ to one of the labels:  
$f(c,d) = l \in \mathcal{L} =$ \{\textit{Supported, Contradictory, Unverified}\}.

\start{Augmenting Labels with Textual Critiques}
In addition to the classification label, we aim to train the evaluator to generate a \textit{textual critique} that explains the judgement, which provides more informative feedback such as which part of the document supports or contradicts the claim. We will leverage such feedback to revise generator responses and use them to train a more factual generator (see \S\ref{sec:improve_factuality} for details).
Formally, we aim to learn the mapping $f: \mathcal{C} \times \mathcal{D} \rightarrow \mathcal{R} \times \mathcal{L}$, which maps each claim-document pair $(c,d)$ to both the textual critique $r \in \mathcal{R}$ and the label $l \in \mathcal{L}$.

As shown in \autoref{fig:evaluator_train}, we prompt an instruction-tuned model $\mathcal{M}$ (e.g., Llama3-70B-chat) to generate both the critique $r_\mathcal{M}$ and label $l_\mathcal{M}$ for whether a claim $c$ is supported. The critique and label are likely to be consistent because the label is generated conditioned on the critique. As a result, if the predicted label $l_\mathcal{M}$ aligns with the ground truth label $l_{GT}$ in the original dataset, the critique is also likely to be aligned and we hence use both the critique $r_\mathcal{M}$ and label $l_\mathcal{M}$ as the new training target. Otherwise, if the predicted label does not align with the ground truth, we discard the whole example.

\start{Augmenting Source Documents using Tools}
To judge the factuality of an arbitrary model-generated claim, we could potentially benefit from \textit{a multiplicity of tools} such as search engines or online knowledge bases. However, the source documents in existing judgment datasets typically come from very restricted sources, such as news corpora or Wikipedia.
To bridge this gap, we obtain additional source documents for each claim by calling the following tools: a search engine (Bing Search API), knowledge base (Wikipedia), and knowledge graph (Google Knowledge Graph API).

As shown in \autoref{fig:evaluator_train}, given the claim $c$ in the original dataset, we prompt an instruction-tuned model to call multiple tools to verify the factuality of the claim (i.e., by generating tool calls such as search queries).
Then we rerank the returned results to obtain a combination of tool-extracted documents $d_t$. 
Similar to critique generation, we prompt the instruction-tuned model $\mathcal{M}$ to predict whether claim $c$ is supported by the tool-extracted documents $d_t$. We add $d_t$ to the train set if the predicted label $f_\mathcal{M}(c, d_t)$ is the same as the ground truth label $f_{GT}(c, d)$ in the original dataset.

The intuition is that if a claim can be supported by some documents, it is likely that we can obtain other supporting sources by calling the tools.
If a claim is hallucinated, it is very unlikely to find any knowledge that supports it with any tools. In both cases, we have $f_{GT}(c, d_t) = f_{GT}(c, d)$.

\start{Training Objective}
After obtaining the augmented training data $\mathcal{TR}_{Eval}=\{(c,d),(r,l)\}$, where each example contains (claim $c$, source document $d$, critique $r$, label $l$), we initialize the evaluator $\mathcal{E}$ with a instruction-tuned model and train it with a standard conditional language modeling objective, maximizing likelihood:
\begin{equation}
    \max _{\mathcal{E}} \mathbb{E}_{(c,d),(r,l) \sim \mathcal{TR}_{Eval}} \log \mathbb{P}_{\mathcal{E}}(r, l \mid c, d).
\label{eq:evaluator_sft}
\end{equation}
\downv
\begin{figure*}[t]
\centering
    \centering
    \includegraphics[width=\linewidth]{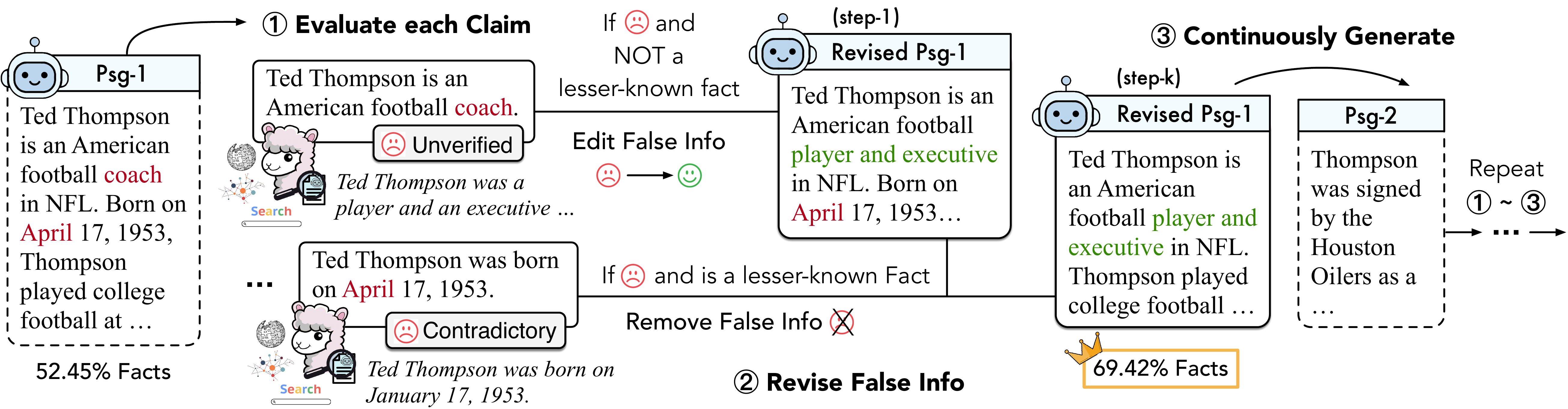}
    \upv
    \vspace{-0.2cm}
    \caption{
    The framework to revise model responses without introducing lesser-known facts. 
    We iteratively 
    (1) use \modelname to evaluate the factuality of each claim, 
    (2) replace false information (if any) with the correct fact or remove it from the response, depending on whether it corresponds to a lesser-known fact,
    and (3) continue generating the next passage. 
    For every claim, we prompt the generator ``\textit{Is this claim factual?}'' without providing any source documents. If the generator outputs \textit{``unknown''}, we assume that the claim corresponds to a lesser-known fact.
    }
    \label{fig:revise}
    \downv
\end{figure*}

\subsection{Improving Generator Factuality with \modelname}
\label{sec:improve_factuality}
In this section, we use our evaluator, \modelname, to improve a generator model's factuality, where we construct training data by revising and scoring the generator's own responses.
Compared to directly training the generator on factuality datasets, our method only requires a prompt set as inputs and hence enjoys much better scalability.

\start{Overview}
As shown in \autoref{fig:overview}, given a prompt, we use the generator to generate $N$ candidate responses.
Then we improve the factuality of each response by using \modelname to evaluate the factuality of each piece of generated information and editing or removing the false information, depending on whether the corresponding fact is rare.
Finally, we use \modelname again to score all original and revised responses and construct training data.

\start{Response Revision}
We aim to improve the factuality of the generator's responses \textit{without introducing lesser-known facts} to training data. The motivation is that as shown in recent research~\citep{understand-finetuning,finetuning-on-new}, forcing the model to generate lesser-known facts that are poorly memorized during pretraining will blur the boundary with memorized facts and other plausible sounding information, which may lead to even more hallucination.
As shown in \autoref{fig:revise}, we iteratively revise each passage with the following three steps: 

\quad [\underline{\textbf{Step}-\circled{1} \textbf{Evaluate}}] 
we prompt an instruction-tuned model (e.g., Llama3-70B-chat) to decompose each response into claims. Then for each claim, we call tools to obtain related documents and apply \modelname to evaluate its factuality with critique. 

\quad [\underline{\textbf{Step}-\circled{2} \textbf{Revise}}] 
If there are any claims that are judged as \textit{``unverified''} or \textit{``contradictory,''} we further check whether it corresponds to a lesser-known fact. 
Specifically, we prompt the generator \textit{``Is this claim factual?''} and output \textit{``true,''} \textit{``false,''} or \textit{``unknown,''} without providing it any external knowledge. We regard the claim as a lesser-known fact if the generator outputs \textit{``unknown.''}

If the claim does not correspond to a lesser-known fact, we prompt the generator to correct the false information based on the critique generated by \modelname. Otherwise, we prompt the generator to remove the false information from the passage.

\quad [\underline{\textbf{Step}-\circled{3} \textbf{Generate}}] 
To reduce error propagation, we use the revised passages as the prefix and continuously generate the next passage.

\start{Generator Training}
We use \modelname to score each original and revised response by computing the percentage of factual claims. Then we train the generator with first supervised finetuning (SFT) and then direct preference optimization (DPO)~\citep{DPO}.
In the SFT stage, we train the generator with the top-$k$ responses as targets, where the responses are ranked by the percentage of factual claims, optimizing with the conditional language modeling objective similar to \autoref{eq:evaluator_sft}.

In the DPO stage~\citep{DPO}, we construct preference data $\mathcal{TR}_{Gen}= \{x, y_w, y_l\}$ as follows: for each prompt $x$, we choose the preferred response $y_w$ from the top-$k$ responses and choose any responses with lower scores than $y_w$ as the rejected response $y_l$.
Suppose we have $N$ original and $N$ revised responses, this gives us $\binom{2N}{2} - \binom{2N-k}{2}$ preference pairs.
We initialize the generator from the SFT checkpoint and optimize the following classification loss:
\begin{equation}
\begin{aligned}
    \max _{\mathcal{G}} 
    \mathbb{E}_{\left(x, y_w, y_l\right) \sim \mathcal{TR}_{Gen}}
    & \left[\right.
    \log \sigma
    \left(\right. \beta 
    \log \frac{\pi_\mathcal{G}\left(y_w \mid x\right)}{\pi_{\text {ref}}\left(y_w \mid x\right)} \\
    & -\beta 
    \log \frac{\pi_\mathcal{G}\left(y_l \mid x\right)}{\pi_{\text {ref}}\left(y_l \mid x\right)}
    \left.\right)
    \left.\right],
\end{aligned}
\end{equation}

where $\sigma$ is the Sigmoid function. The reference policy $\pi_{\text {ref}}$ is computed by the SFT checkpoint.

\begin{table*}[t]
\centering
\upv
\resizebox{\textwidth}{!}{
\begin{tabular}{lccccccccccc}
\toprule
& \multicolumn{10}{c}{\textbf{LLM-AggreFact} (\emph{without} threshold tuning)} & \\
\cmidrule(r){2-11} 

\multirow{2}{*}{\bf \quad Model Name} &
  \multicolumn{2}{c}{\textbf{\textsc{AggreFact}}} &
  \multicolumn{2}{c}{\textbf{\textsc{TofuEval}}} &
  \multirow{2}{*}{\textbf{\textsc{Wice}}} &
  \multirow{2}{*}{\textbf{\textsc{Reveal}}} &
  \multirow{2}{*}{\begin{tabular}[c]{@{}c@{}}\textbf{\textsc{Claim}}\\ \textbf{\textsc{Verify}}\end{tabular}} &
  \multirow{2}{*}{\begin{tabular}[c]{@{}c@{}}\textbf{\textsc{Fact}}\\ \textbf{\textsc{Check}}\end{tabular}} &
  \multirow{2}{*}{\begin{tabular}[c]{@{}c@{}}\textbf{\textsc{Expert}}\\ \textbf{\textsc{QA}}\end{tabular}} &
  \multirow{2}{*}{ \textbf{\textsc{Lfqa}}} &
  \multirow{2}{*}{\textbf{Avg}} \\
\cmidrule(r){2-3} \cmidrule(r){4-5} 
& \textbf{CNN}  & \textbf{XSum} & \textbf{MediaS} & \textbf{MeetB} & & & & & & & \\
\midrule
\dlrow \multicolumn{10}{l}{\textit{Open-source Models (Llama3-8B based)}} & & \multicolumn{1}{|c}{} \\
\quad Llama3-8B-chat & 50.9 & 58.2 & 63.9 & 72.4 & 65.1 & 85.2 & 63.8 & 76.5 & 55.9 & 72.3 & \multicolumn{1}{|c}{66.4} \\
\quad \modelname \textit{(Vanilla SFT)} & \underline{63.2} & 73.4 & 66.6 & 77.7 & 64.6 & 86.4 & 72.5 & 73.0 & 57.9 & 83.1 & \multicolumn{1}{|c}{71.8} \\
\quad \modelname \textit{(Critique Only)} & 59.5 & \underline{74.7} & 68.4 & 80.0 & 71.7 & 88.0 & \underline{74.3} & 74.2 & 59.6 & \textbf{87.0} & \multicolumn{1}{|c}{\underline{73.7}} \\
\quad \modelname \textit{(Full)} & 62.1 & 72.4 & \textbf{70.9} & \textbf{80.3} & \underline{76.0} & \textbf{88.6} & \textbf{74.9} & 74.4 & \textbf{60.3} & \underline{86.9} & \textbf{74.7} \\
\midrule 
\dlrow \multicolumn{10}{l}{\textit{Other Open-source Models (47B-123B)}} & & \multicolumn{1}{|c}{} \\
\quad Mistral-8x7B$\dagger$ & 55.0 & 65.5 & \underline{68.5} & 73.3 & 63.8 & 80.8 & 64.3 & 75.1 & 56.3 & 70.8 & \multicolumn{1}{|c}{67.3}  \\
\quad Llama3-70B-chat & \bf 63.3 & 71.3 & 67.9 & 75.2 & 74.8 & 86.7 & 67.3 & \underline{78.3} & 58.4 & 82.9 & \multicolumn{1}{|c}{72.6} \\
\quad Mistral-123B$\dagger$ & 58.4 & \textbf{76.3} & 67.3 & \underline{78.9} & \bf 76.6 & \underline{88.4} & 67.6 & \bf 79.0 & \underline{60.0} & 81.7 & 73.4 \\
\midrule
\midrule 
\dlrow \multicolumn{10}{l}{\textit{Proprietary Models or Distilled Models}} & & \\
\quad Gemini-Pro$\dagger$ & 49.4 & 60.6 & 63.8 & 65.8 & 65.8 & 85.5 & 61.8 & 76.8 & 56.8 & 75.9 & \multicolumn{1}{|c}{66.2} \\
\quad GPT-3.5$\dagger$ & 63.2 & 72.4 & 66.8 & 73.4 & 68.5 & 84.7 & 65.2 & 70.8 & 57.2 & 73.8 & \multicolumn{1}{|c}{69.6} \\
\quad Claude-2.1$\dagger$ & 59.9 & 66.4 & 69.2 & 72.3 & 64.3 & 88.2 & 69.7 & 79.3 & 59.8 & 78.2 & \multicolumn{1}{|c}{70.7} \\
\quad Claude-3 Opus$\dagger$ & 65.2 & 72.4 & 74.1 & 82.4 & 75.0 & 83.8 & 69.3 & 78.8 & 58.8 & 81.6 & \multicolumn{1}{|c}{74.1} \\
\quad MiniCheck-FT5$\dagger$ & 69.9	& 74.3 & 73.6 & 77.3 & 72.2 & 86.2 & 74.6 & 74.7 & 59.0 & 85.2 & \multicolumn{1}{|c}{74.7} \\
\quad GPT-4$\dagger$ & 66.7 & 76.5 & 71.4 & 79.9 & 80.4 & 87.8 & 67.6 & 79.9 & 59.2 & 83.1 & \multicolumn{1}{|c}{75.3} \\

\bottomrule
\end{tabular}
}
\caption{Performance (BAcc) of evaluator models on the test split of LLM-AggreFact. We separate Llama3-8B-chat based models, larger open-source models, and proprietary models into different blocks. We highlight the best-performing open-source model for each dataset.
Results with $\dagger$ are reported in \citet{llm-aggrefact}.
} 
\label{tab:res-llm-aggrefact}
\end{table*}

\section{Experiments on Evaluator Training}
In this section, we aim to answer the research question: (\textbf{RQ1}) Can \modelname correctly judge the factuality of model-generated claims?

\subsection{Experimental Setup}
\start{Training Details}
We initialize \modelname from Llama3-8B-chat and train it on a set of public factuality judgment datasets which \citet{FLAMe} is trained on.
To ensure label accuracy, we focus on datasets with human judgment on model responses, including summarization datasets: XSum Hallucination~\citep{xsum-hallucination}, QAGS~\citep{QAGS}, FRANK~\citep{FRANK}, question-answering datasets: RAGTruth~\citep{ragtruth}, FActScore~\citep{factscore}, and dialogue datasets: Q$^2$~\citep{Q2}, FaithDial~\citep{faithdial}, BEGIN~\citep{begin}.
We provide implementation and dataset details in \S\ref{sec:implementation_details} and \S\ref{sec:data_details}.

\start{Evaluation Dataset and Metric}
We evaluate the evaluators on the LLM-AggreFact benchmark~\citep{llm-aggrefact}, a combination of 10 datasets covering three tasks: fact verification, summarization, and long-form QA.
All datasets contain human-annotated (document, claim, label) tuples. 

We follow \citet{llm-aggrefact} and use balanced accuracy (BAcc) as the evaluation metric: $\text{BAcc} = \frac{1}{2}\left(\frac{\mathrm{TP}}{\mathrm{TP}+\mathrm{FN}}+\frac{\mathrm{TN}}{\mathrm{TN}+\mathrm{FP}}\right)$, where $\mathrm{TP}$, $\mathrm{TN}$, $\mathrm{FP}$, and $\mathrm{FN}$ represent true/false positives/negatives. 

\start{Baselines and Ablations}
We compare to the LLM-based fact-checkers reported by \citet{llm-aggrefact}.
We do not include results reported in \citet{FLAMe} because they use a different metric.

In addition, we compare to two ablations: (1) \modelname (\textit{Vanilla SFT}), which is trained on the original public datasets with no augmentation, and (2) \modelname (\textit{Critique Only}), where we only generate textual critiques and not source documents.

\subsection{Main Results}
As shown in \autoref{tab:res-llm-aggrefact}, after training on our augmented datasets, \modelname improves Llama3-8B-chat by 8.3\% BAcc and outperforms all the open-source LLMs, including models with significantly more parameters (e.g., Mistral-123B). It also outperforms strong proprietary models such as Claude-3 Opus.
When compared with its ablations, \modelname consistently outperforms the Vanilla SFT model on 8 out of 10 datasets, with average gain of 2.9\% BAcc.
The performance of \modelname (\textit{Critique Only}) is between Vanilla SFT and the full model, which indicates the utility of both augmentation methods.

Among all the datasets, we observe that the performance on Wice decreases after Vanilla SFT, but is largely improved by \modelname. 
One possible reason is that some training datasets such as Q$^2$ contain claims that are labeled as ``factual'' but are only partially supported by the documents.
We hypothesize that such noisy examples strongly affect the performance on Wice, which contains as high as 54.7\% of such ``partially supported'' examples.
However, by filtering out examples where Llama3-70B-chat cannot generate explanations, we filter out a large percent of such noisy examples and hence improve the final judgment accuracy.

\subsection{Result Analyses}
\start{Accuracy of Augmented Critique and Source Documents}
With our data augmentation methods, we equipped 77.2\% of the training examples with textual critiques, and generate new source documents for 54.1\% of the examples with the combination of our three tools. 
To verify the data quality, we randomly sampled 45 examples where we successfully obtain critiques or source documents (15 for each label) and manually inspect the accuracy.
Specifically, we check whether both the critique and the label correctly reflect the relationship between the claim and the source documents.

As shown in \autoref{tab:aug-acc}, 95.6\% of the augmented critiques and 97.8\% of the tool-extracted documents are accurate. 
Although in one of the wrongly-labeled examples, our method mislabels ``unverified'' as ``contradictory'', it does not affect the final conclusion that the claim is not factual.

\begin{table}[t]
\centering
\resizebox{\linewidth}{!}{
\begin{tabular}{lccc}
\toprule
\bf Pred Label ($\rightarrow$) & \bf Supported & \bf Contradictory & \bf Unverified \\ 
\midrule
Critique Acc & 15/15 & 14/15 & 14/15 \\ 
Tool-ext Doc Acc & 14/15 & 15/15 & 15/15 \\
\bottomrule
\end{tabular}
}
\caption{
The accuracy of the critiques and the tool-extracted source documents we obtained. We randomly sampled 15 examples for each predicted label and manually check the accuracy of each example.
}
\label{tab:aug-acc}
\end{table}

\start{Case Studies}
We further show three concrete examples of our augmented critiques and source documents.
In the first example (\autoref{tab:case_1}), our generated critique correctly explain the judgment label: the claim mentions a call for a national project, which does not appear in the documents.

In the second example (\autoref{tab:case_2}), the original document is a CNN news report. By calling the search engine, we obtain other news articles written by diverse news agencies on the same event (i.e., Kenneth Morgan's murder). Such tool-extracted documents increase the diversity of documents in the train set, while still having high label accuracy.

In the third example (\autoref{tab:case_3}), we obtain Chadwick Boseman's birthday information from all three tools: knowledge graph, Wikipedia, an search engine, where the knowledge graph provides more structured and concise knowledge, while Wikipedia returns a long paragraph containing the information.
Compared to the original documents, our tool-extracted documents have more diverse formats, improving the evaluator's generalizability.

\section{Experiments on Generator Factuality}
We aim to answer the research questions: 
(\textbf{RQ2}) Can we leverage \modelname to improve the generator's factuality?
(\textbf{RQ3}) How well can our training recipe improve the generator's factuality?

\subsection{Experimental Setup}
\start{Datasets and Evaluation Metric}
Following prior works~\citep{EVER,self-eval-skt}, we conduct experiments on FActScore~\citep{factscore} and TruthfulQA~\citep{truthfulqa}.
For FActScore, we randomly split the unlabeled split into 400 training and 100 test prompts.
We compute the ``\% Facts'' metric by extracting correct and incorrect facts in each response, where we use Llama3-70B-chat to decompose the responses and use the same three tools in training to obtain source documents.
For TruthfulQA, we randomly select 3 examples from each of the 38 categories as the training set and use the remaining 703 examples as the test set.
We follow the ``generation'' setting and use the finetuned evaluator in the original paper to compute ``\% True*Info'', the percentage of responses that are both truthful and informative.

\begin{table}[t]
\centering
\resizebox{\linewidth}{!}{
\begin{tabular}{lcccc}
\toprule
\multirow{2}{*}{\bf Model Name} 
& \multicolumn{3}{c}{\bf FActScore} 
& \bf TruthfulQA \\
\cmidrule(lr){2-4}
\cmidrule(lr){5-5}
& \bf \# Facts & \bf \# Errors & \bf \% Facts 
& \bf \% True*Info \\
\midrule
Llama2-7B-chat & 10.70 & 17.04 & 38.57 & 38.83 \\
+ SFT & 10.76 & 15.59 & 40.83 & 45.52 \\
+ Self-Eval-SKT & 11.02 & 14.18 & 43.73 & 48.65 \\
+ EVER-Pref & \bf 11.24 & 15.11 & 42.66 & 51.07 \\
+ FactTune-FS & 11.23 & 12.87 & 46.60 & 52.48 \\
\hdashline
\textit{\colorgrey{(Our Method)}} & & & & \\
+ E/R + Coarse & 10.84 & \bf 8.72 & \bf 55.43 & \bf 56.47 \\
\midrule
Llama3-8B-chat & 17.83 & 17.16 & 50.96 & 58.89 \\
+ SFT & 20.05 & 18.13 & 52.52 & 59.17 \\
+ Self-Eval-SKT & 18.69 & 14.22	& 56.80 & 61.88 \\
+ EVER-Pref & 20.25 & 15.16 & 57.18 & 63.01 \\
+ FactTune-FS & 18.77 & 13.34 & 58.45 & 64.58 \\
\hdashline
\textit{\colorgrey{(Our Method)}} & & & & \\
+ E/R + Coarse & \bf 20.40 & \bf 10.79 & \bf 65.41 & \bf 67.14 \\
\bottomrule
\end{tabular}
}
\caption{
Comparison between our method and baselines on FActScore and TruthfulQA. 
``E/R'' stands for ``Edit/Remove''.
All baselines use the zero-shot models as the evaluator in training and our method uses \modelname.
}
\label{tab:res-factscore}
\end{table}

\begin{table}[t]
\centering
\resizebox{0.9\linewidth}{!}{
\begin{tabular}{lccc}
\toprule
\multirow{2}{*}{\bf Model Name} 
& \multicolumn{3}{c}{\bf FActScore} \\
\cmidrule(lr){2-4}
& \bf \# Facts & \bf \# Errors & \bf \% Facts \\
\midrule
\multicolumn{4}{l}{\textit{\colorgrey{(Ablations with \modelname as the evaluator)}}} \\
Llama3-8B-chat & 17.83 & 17.16 & 50.96 \\
+ SFT + \modelname & 21.19 & 16.47 & 56.26 \\
+ Edit & \bf 20.68 & 14.42 & 58.91 \\
+ Coarse & 20.07 & 12.89 & 60.89 \\
+ Edit + Coarse & 20.03 & 11.09 & 64.37 \\
\hdashline
\textit{\colorgrey{(Our Full Method)}} & & & \\
+ E/R + Coarse & 20.40 & \bf 10.79 & \bf 65.41 \\
\bottomrule
\end{tabular}
}
\caption{
Ablation study that compares different training recipes when equipped with \modelname as the evaluator.
``SFT + \modelname'', ``Edit'', and ``Coarse'' denote equipping ``SFT'', ``EVER-Pref'', and ``FactTune-FS'' in \autoref{tab:res-factscore} with the \modelname evaluator.
}
\label{tab:res-ablation}
\end{table}

\begin{figure*}[t]
  \centering
  \begin{subfigure}[b]{0.33\textwidth}
    \includegraphics[width=\textwidth]{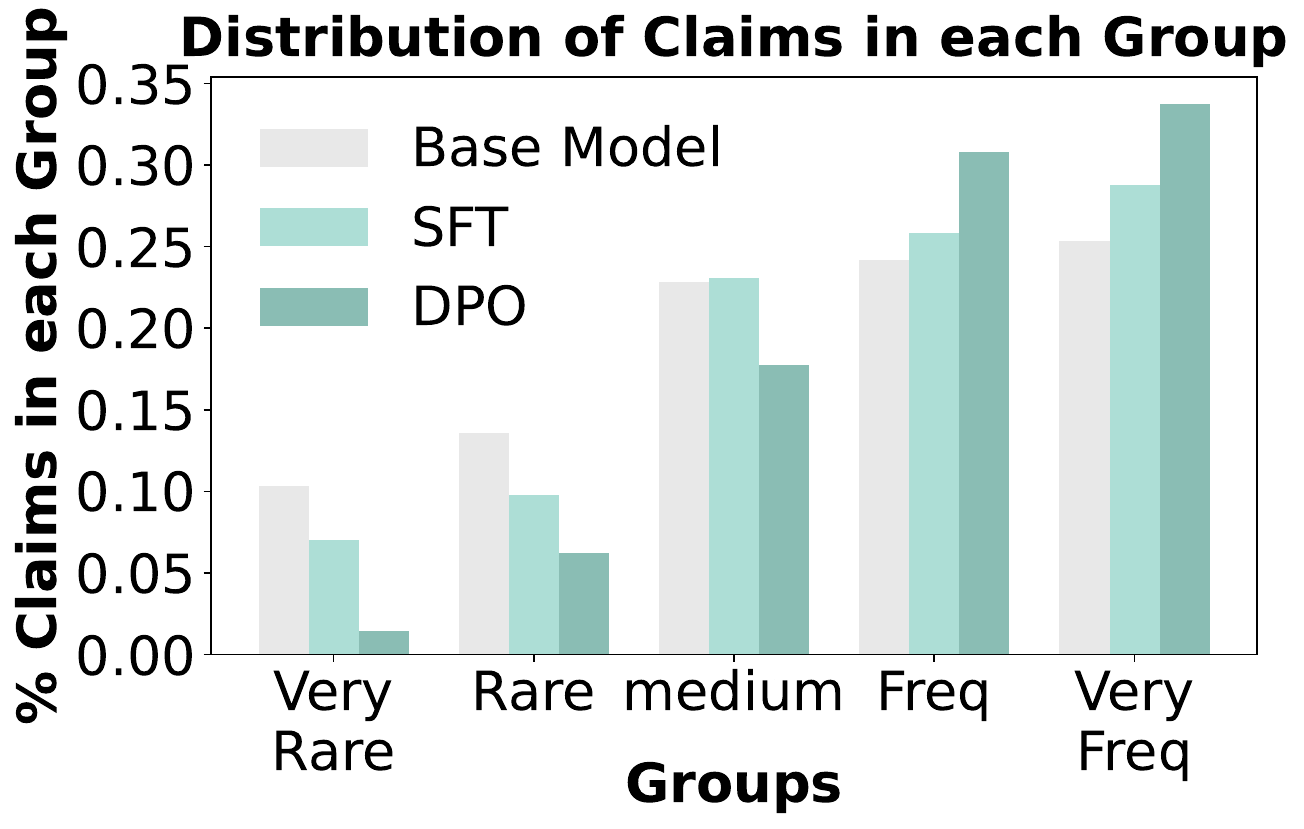}
    \caption{Distribution of claims}
    \label{fig:response_shift_distribution_claims}
  \end{subfigure}
  \begin{subfigure}[b]{0.32\textwidth}
    \includegraphics[width=\textwidth]{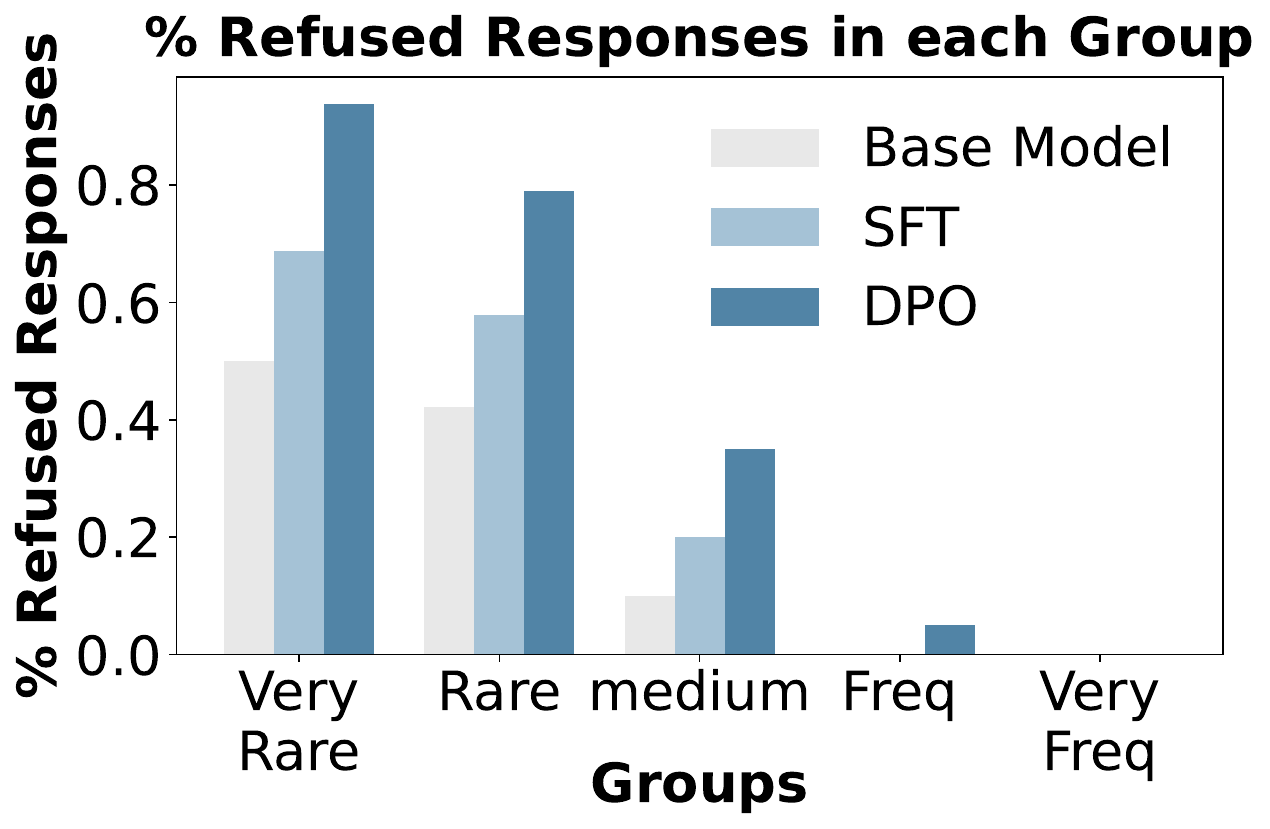}
    \caption{\% Refused responses}
    \label{fig:response_shift_refused_responses}
  \end{subfigure}
  \begin{subfigure}[b]{0.32\textwidth}
    \includegraphics[width=\textwidth]{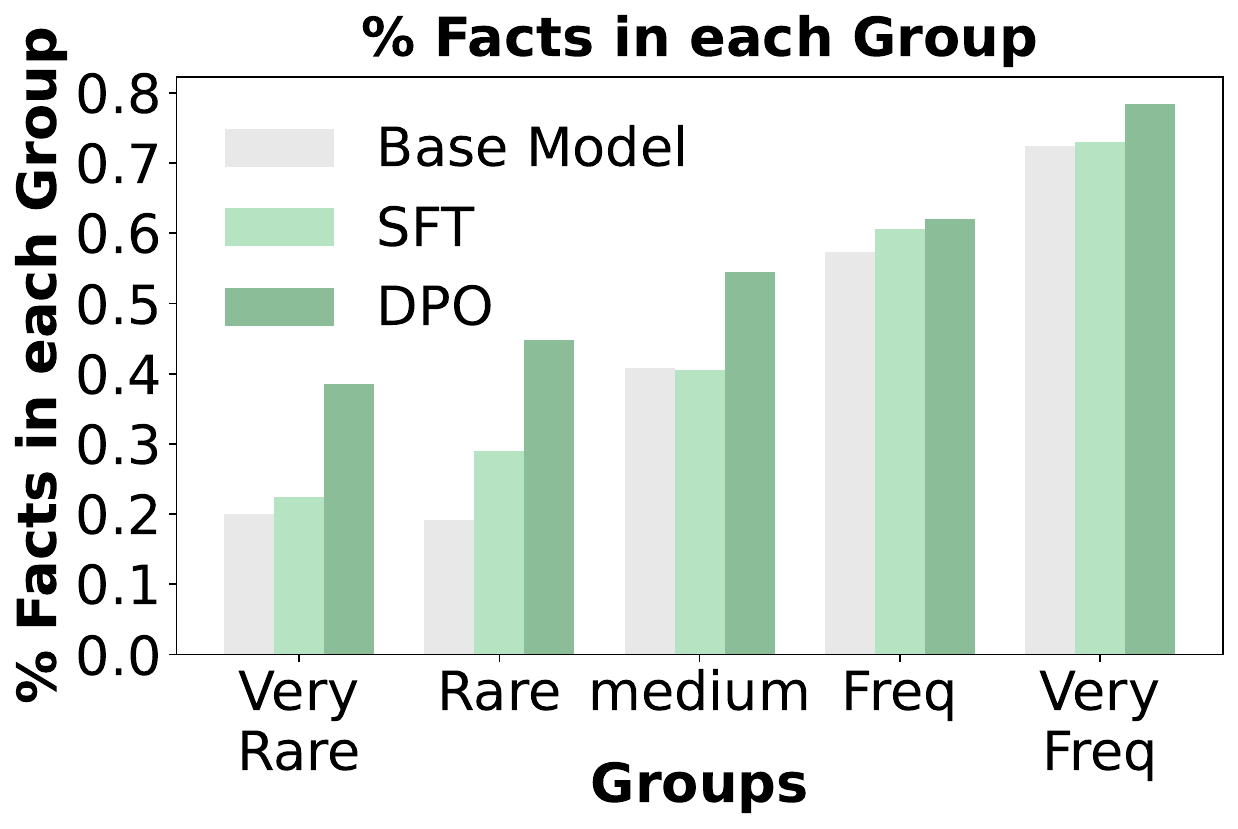}
    \caption{\% Facts}
    \label{fig:response_shift_num_facts}
  \end{subfigure}
  
  \caption{Statistics of generated responses, where we group the prompts by the popularity of the person to write biography for, as labeled in the FActScore dataset. We compare zero-shot Llama3-8B-chat (denoted as \textit{``Base Model''}), SFT + \modelname (\textit{``SFT''}), and our method, Edit/Remove + Coarse (\textit{``DPO''}).}
  \label{fig:response_shift}
\end{figure*}

\begin{figure}[t]
    \centering
    \includegraphics[width=0.95\linewidth]{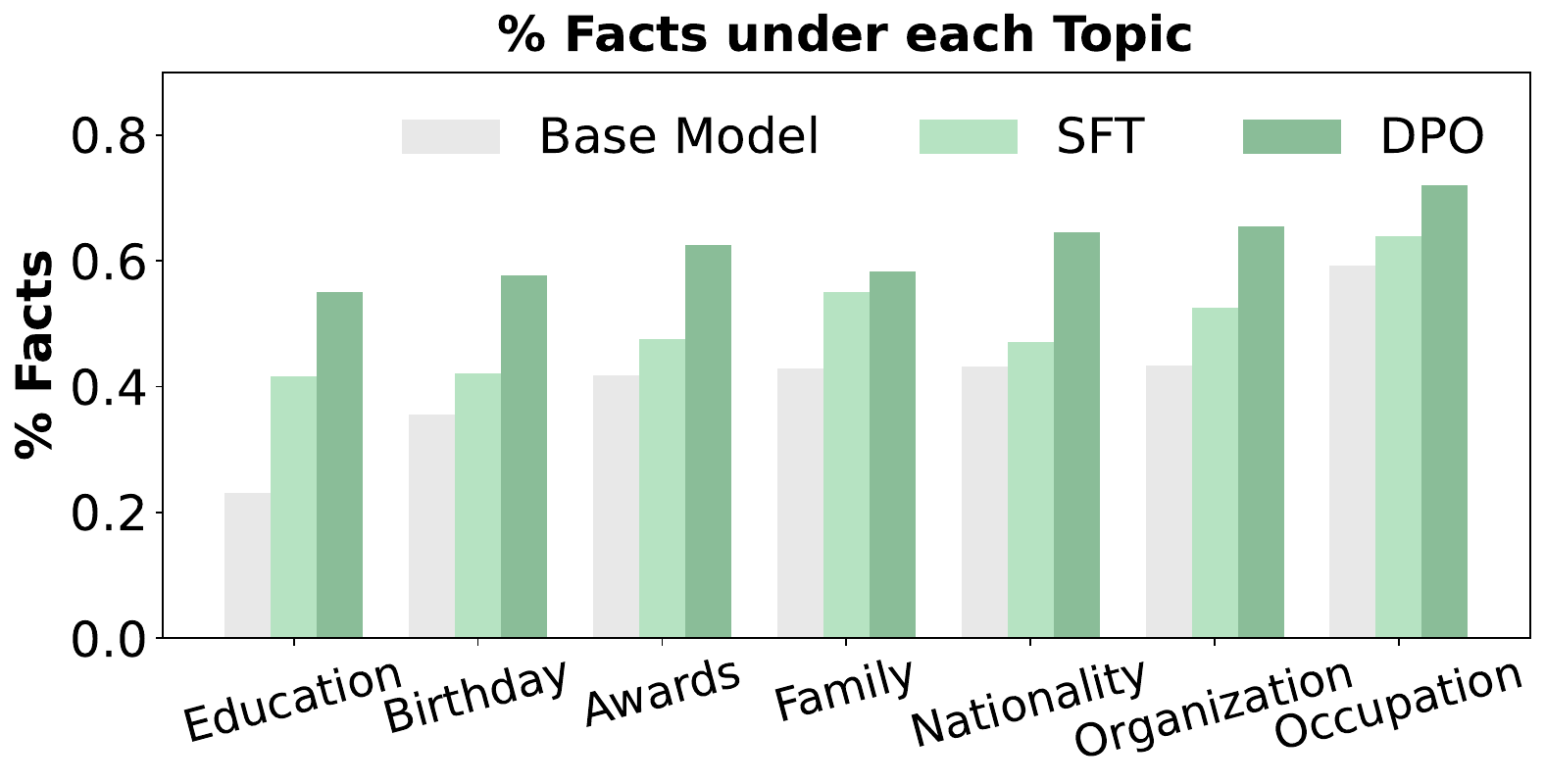}
    \upv
    \caption{The factuality of claims with different topics, where we pre-define the topics and use Llama3-70B-chat to predict whether each claim covers any topic(s).}
    \label{fig:claim_acc_by_cat}
    \downv
\end{figure}

\start{Baselines and Ablations}
We implement four baselines: SFT, FactTune-FS~\citep{FactTune}, Self-Eval-SKT~\cite{self-eval-skt}, and EVER-Pref~\citep{EVER}. 
All methods sample $N$ candidate responses for each training prompt.
SFT first uses the generator itself to score responses by computing the percentage of factual claims and finetunes with the best response.
FactTune-FS first finetunes on all candidates and then uses all $\binom{N}{2}$ candidate pairs as DPO pairs, preferring the one with a higher score based on retrieved context.
Self-Eval-SKT self-trains an evaluator using the model's own knowledge and uses the evaluator to score responses with no external context, also resulting in $\binom{N}{2}$ preference pairs.
EVER-Pref uses all $N$ candidates as rejected responses and constructs a preferred response by iteratively evaluating and correcting false information in each passage.

For ablations, we first equip SFT, FactTune, and EVER with our evaluator, \modelname, and the three external tools, denoted as ``SFT + \modelname'', ``Coarse'', and ``Edit'' in \autoref{tab:res-ablation}, respectively. 
Then we implement ``Edit + Coarse'', which corrects all false information without checking whether it corresponds to a lesser-known fact.
We denote our full method as ``E/R + Coarse'' (Edit/Remove + Coarse).
More implementation details can be found in \S\ref{sec:implementation_details}.

Note that we do not compare to retrieval-augmented (RAG) methods~\cite{RAG} because it is not a fair comparison. RAG-based methods require access to external knowledge sources or tools in inference, while in our experiments, we only used ``\texttt{tell me a bio of <entity>}'' as the prompt. 
Furthermore, while most RAG-based methods aim to improve the understanding of the long retrieved context, our main focus is to train the model to better utilize the knowledge it has seen during pretraining.

We also do not compare to inference-time methods~\cite{ITI,DoLA} because these methods need to call the model multiple times in inference, introducing heavy latency issues. In the real world, such latency is unacceptable in production. In comparison, our method uses the standard random decoding in inference and does not introduce extra inference-time cost. Furthermore, in principle, our method is orthogonal to those inference-time methods and can be combined.

\begin{figure}[t]
  \centering
  \begin{subfigure}[b]{0.245\textwidth}
    \includegraphics[width=\textwidth]{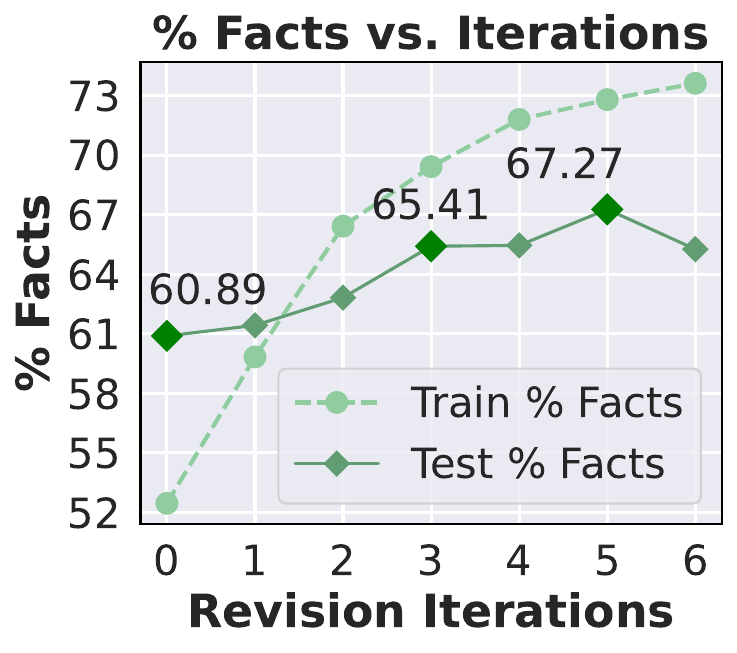}
    \caption{Iterations of Revision}
    \label{fig:hyper_param_iter}
  \end{subfigure}
  \begin{subfigure}[b]{0.23\textwidth}
    \includegraphics[width=\textwidth]{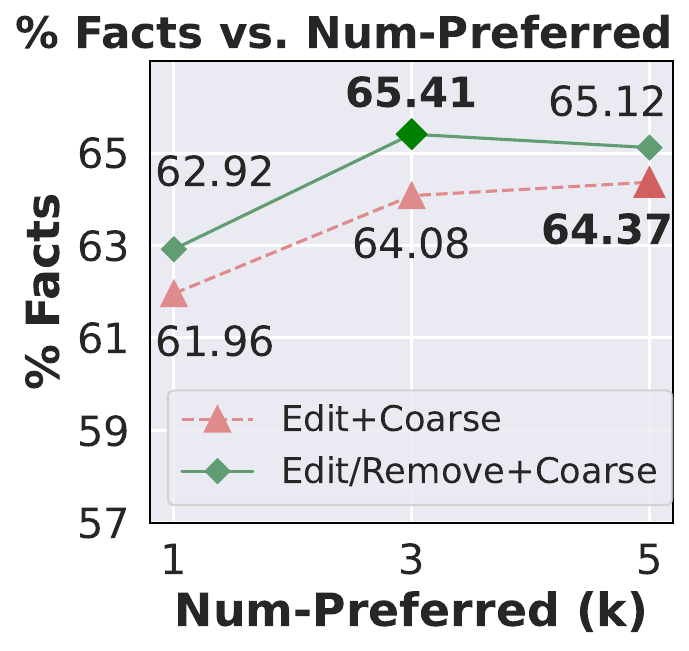}
    \caption{top-$k$}
    \label{fig:hyper_param_k}
  \end{subfigure}
  \caption{Hyper-parameter analysis. We investigate how percentage of facts changes
  (a) as number of iterations to revise candidate responses increases, and 
  (b) with different numbers of preferred responses for each prompt.
  }
  \label{fig:hyper-param}
\end{figure}

\subsection{Main Results}
Results in \autoref{tab:res-factscore} answer (\textbf{RQ2}) and show that our method significantly improves Llama2/Llama3's factuality by 16.85/14.45\% on FActScore and 17.64\%/8.25\% on TruthfulQA.
Results also show that our method significantly outperforms the best baseline on factuality training (e.g., by 8.83/6.96\% on FActScore for Llama2/Llama3).

\autoref{tab:res-ablation} presents the ablations with \modelname as the evaluator.
The comparison between \autoref{tab:res-factscore} and \autoref{tab:res-ablation} shows that \modelname can improve the performance of SFT, EVER, and FactTune. 
For example, on FActScore, with Llama3 as the base model, SFT with \modelname as the evaluator outperforms SFT with Llama3 as the evaluator by 3.74\%.

In \autoref{tab:res-ablation}, we observe that our method of combining response editing/removing with coarse-level scoring achieves significantly better performance.
In particular, our full method, which only corrects false information with common facts, outperforms ``Edit + Coarse'', which could introduce both common and lesser-known facts into the training data.
The above results answer (\textbf{RQ3}), demonstrating the effectiveness of our training recipe.

\subsection{Result Analyses}

\start{Distribution of Generated Claims}
We first group the prompts by the popularity of the person to write biography for, which is provided as meta data in the FActScore dataset, and then compare the distribution of generated claims in each group before and after training.
As shown in \autoref{fig:response_shift_distribution_claims}, after training, the generator outputs less information for unfamiliar people and outputs more information for popular ones, suggesting that it learns to only generate information that is likely to be factual.

In addition, as shown in \autoref{fig:response_shift_refused_responses}, we observe that the generator refuses to generate responses more frequently for rare entities (e.g., by generating ``I apologize, but I'm not familiar with this person.''), and almost never refuses for frequent ones. 
This aligns with previous research's conclusion~\citep{unfamiliar-finetuning} that training the model to say ``I don't know'' to unfamiliar prompts reduces hallucination.

\start{Performance Breakdown}
We further check the generator's performance (i.e., percentage of factual claims) on different groups of prompts. 
As shown in \autoref{fig:response_shift_num_facts}, we observe that our method achieves consistent performance gain over all groups of prompts, which is another reason why our method obtains higher overall performance.

Similarly, we check the performance on claims describing different topics, where we first come up with a list of topics and then prompts Llama3-70B-chat to determine whether each claim covers any of the topics. 
In \autoref{fig:claim_acc_by_cat}, our method generates more factual claims under all the topics, and the performance gain is larger on unfamiliar topics (i.e., topics where the generator has lower scores). 

\start{Hyper-parameter Analysis}
We first alternate the number of iterations to revise the responses and inspect the testing and training accuracy (i.e., the average \% of facts of the best preferred responses).
As shown in \autoref{fig:hyper_param_iter}, with more revision iterations, we can always obtain preferred responses with better factuality, but the test performance converges after the third iteration. 
In other words, training data with fewer factual errors does not always transfer to better test performance.

In our experiments, we only choose the top-$k$ candidates (ranked by the percentage of facts) as preferred responses. We investigate the effect of $k$ in \autoref{fig:hyper_param_k}. We observe that training on top-3 and top-5 responses leads to similar performance. With all the $k$s in our experiments, our method consistently outperforms the ``Edit + Coarse'' ablation.

\section{Related Work}
\start{Factuality Evaluation} To judge the factuality of long-form model responses, recent works have presented fine-grained level evaluation frameworks that judge each piece of generated fact individually~\citep{factscore,factool,doclens,longform}.
Such frameworks generally leverage an LM evaluator to make judgments.

To train LM evaluators, one line of work constructs new training datasets by collecting human judgement~\citep{alpaca_eval,TIGERScore}. 
Another line of work distills open-source evaluators from proprietary models such as GPT-4, training the evaluator to generate textual critique~\citep{shepherd,ultrafeedback,AutoJ} or fine-grained judgment~\citep{prometheus,FAVA}.
A recent work \citep{FLAMe} leverages a combination of existing public datasets to train an evaluator.
We further augment the public datasets with more diverse knowledge sources and more informative judgment feedback.

\start{Enhancing Generator's Factuality}
To reduce hallucinations, previous works present inference-time methods, including re-computing token probabilities~\cite{CAD,DoLA} or conducting post-editing~\cite{self-critiquing,self-improve,self-correct,self-refine,CRITIC,FAVA}. Such methods inevitably suffer from latency issues.

Another approach is to train the model for factuality. Following general reward modeling methods that produce a score for the entire response~\citep{RLHF,InstructGPT,RLHF-QA}, FactTune \citep{FactTune} trains the generator by preferring responses with higher percentage of facts, which is limited by the generator's capability. 
EVER~\citep{EVER} constructs high-quality responses by correcting false information, which may introduce lesser-known facts and could potentially harm model factuality~\citep{understand-finetuning}.
Both methods either prompt proprietary model or use the generator itself to evaluate its own factuality, which are either restricted by terms of use or suffer from self-bias.

Our method is different from existing methods in the following aspects: (1) we combine response revision and scoring to construct high-quality responses while ensuring the correctness of preference ranking, (2) we only correct false information with common facts and remove other misinformation from training, and (3) we use our open-source evaluator, \modelname, for both revision and scoring.

\vspace{0.35cm}
\section{Conclusions}
We improve LM generators' factuality by training an open-source evaluator model, \modelname.
To train \modelname to leverage diverse knowledge sources and to generate more informative feedback, 
we equip a combination of public datasets with textual critique along with judgment scores and obtain additional source documents by calling a multiplicity of tools.
We then present a training recipe that leverages \modelname to finetune LM generators for better factuality, where we construct preference data by prompting \modelname to revise and score the generator's responses, without introducing lesser-known facts in training.
Experiments show that \modelname outperforms strong proprietary models such as Claude-3 on LLM-AggreFact.
Our factuality training method improves Llama3-8B-chat's factuality performance by 14.45\% on FActScore and 8.25\% on TruthfulQA.

\section{Limitations}
We list the limitations of this work as follows:

\start{Evaluator Training Data} We only train our evaluator on human-annotated datasets on model responses. We have not investigate the effect of other datasets (e.g., synthetic datasets or human-written claim datasets) and we leave it to future works. 

Furthermore, in this work, we focus on text-to-text generation and do not train our evaluator on math reasoning or programming tasks, which are also classified as factuality tasks in some existing works~\citep{factool}.

\start{Experiments on Improving Factuality} To test our generator's performance, we following existing works~\citep{FactTune,EVER} and only experiment on one public dataset: FActScore. In principle, one can also apply our training recipe to other prompt datasets.

\section{Acknowledgement}
We thank Xinran Zhao, Tim Franzmeyer, Zichun Yu, and Pengfei Liu for their helpful feedback on this work.
This work was supported in part by NSF grant DSES-2222762.
Yiqing Xie is supported by the Carnegie Mellon University Presidential Fellowship in the Language Technologies Institute.

\bibliography{custom}

\begin{thebibliography}{52}
\providecommand{\natexlab}[1]{#1}

\bibitem[{Bajaj et~al.(2018)Bajaj, Campos, Craswell, Deng, Gao, Liu, Majumder,
  McNamara, Mitra, Nguyen, Rosenberg, Song, Stoica, Tiwary, and Wang}]{msmarco}
Payal Bajaj, Daniel Campos, Nick Craswell, Li~Deng, Jianfeng Gao, Xiaodong Liu,
  Rangan Majumder, Andrew McNamara, Bhaskar Mitra, Tri Nguyen, Mir Rosenberg,
  Xia Song, Alina Stoica, Saurabh Tiwary, and Tong Wang. 2018.
\newblock \href {https://arxiv.org/abs/1611.09268} {Ms marco: A human generated
  machine reading comprehension dataset}.
\newblock \emph{Preprint}, arXiv:1611.09268.

\bibitem[{Chern et~al.(2023)Chern, Chern, Chen, Yuan, Feng, Zhou, He, Neubig,
  and Liu}]{factool}
I-Chun Chern, Steffi Chern, Shiqi Chen, Weizhe Yuan, Kehua Feng, Chunting Zhou,
  Junxian He, Graham Neubig, and Pengfei Liu. 2023.
\newblock \href {https://arxiv.org/abs/2307.13528} {Factool: Factuality
  detection in generative ai -- a tool augmented framework for multi-task and
  multi-domain scenarios}.
\newblock \emph{Preprint}, arXiv:2307.13528.

\bibitem[{Chuang et~al.(2024)Chuang, Xie, Luo, Kim, Glass, and He}]{DoLA}
Yung-Sung Chuang, Yujia Xie, Hongyin Luo, Yoon Kim, James~R. Glass, and
  Pengcheng He. 2024.
\newblock \href {https://openreview.net/forum?id=Th6NyL07na} {Dola: Decoding by
  contrasting layers improves factuality in large language models}.
\newblock In \emph{The Twelfth International Conference on Learning
  Representations}.

\bibitem[{Cui et~al.(2024)Cui, Yuan, Ding, Yao, He, Zhu, Ni, Xie, Xie, Lin,
  Liu, and Sun}]{ultrafeedback}
Ganqu Cui, Lifan Yuan, Ning Ding, Guanming Yao, Bingxiang He, Wei Zhu, Yuan Ni,
  Guotong Xie, Ruobing Xie, Yankai Lin, Zhiyuan Liu, and Maosong Sun. 2024.
\newblock \href {https://arxiv.org/abs/2310.01377} {Ultrafeedback: Boosting
  language models with scaled ai feedback}.
\newblock \emph{Preprint}, arXiv:2310.01377.

\bibitem[{Dinan et~al.(2019)Dinan, Roller, Shuster, Fan, Auli, and
  Weston}]{wow}
Emily Dinan, Stephen Roller, Kurt Shuster, Angela Fan, Michael Auli, and Jason
  Weston. 2019.
\newblock \href {https://arxiv.org/abs/1811.01241} {Wizard of wikipedia:
  Knowledge-powered conversational agents}.
\newblock \emph{Preprint}, arXiv:1811.01241.

\bibitem[{Dziri et~al.(2022{\natexlab{a}})Dziri, Kamalloo, Milton, Zaiane, Yu,
  Ponti, and Reddy}]{faithdial}
Nouha Dziri, Ehsan Kamalloo, Sivan Milton, Osmar Zaiane, Mo~Yu, Edoardo~M.
  Ponti, and Siva Reddy. 2022{\natexlab{a}}.
\newblock \href {https://doi.org/10.1162/tacl_a_00529} {{F}aith{D}ial: A
  faithful benchmark for information-seeking dialogue}.
\newblock \emph{Transactions of the Association for Computational Linguistics},
  10.

\bibitem[{Dziri et~al.(2022{\natexlab{b}})Dziri, Rashkin, Linzen, and
  Reitter}]{begin}
Nouha Dziri, Hannah Rashkin, Tal Linzen, and David Reitter. 2022{\natexlab{b}}.
\newblock \href {https://doi.org/10.1162/tacl_a_00506} {Evaluating attribution
  in dialogue systems: The {BEGIN} benchmark}.
\newblock \emph{Transactions of the Association for Computational Linguistics},
  10.

\bibitem[{Gekhman et~al.(2024)Gekhman, Yona, Aharoni, Eyal, Feder, Reichart,
  and Herzig}]{finetuning-on-new}
Zorik Gekhman, Gal Yona, Roee Aharoni, Matan Eyal, Amir Feder, Roi Reichart,
  and Jonathan Herzig. 2024.
\newblock \href {https://arxiv.org/abs/2405.05904} {Does fine-tuning llms on
  new knowledge encourage hallucinations?}
\newblock \emph{Preprint}, arXiv:2405.05904.

\bibitem[{Ghosal et~al.(2024)Ghosal, Hashimoto, and
  Raghunathan}]{understand-finetuning}
Gaurav~Rohit Ghosal, Tatsunori Hashimoto, and Aditi Raghunathan. 2024.
\newblock \href {https://openreview.net/forum?id=cPsn9AcOYh} {Understanding
  finetuning for factual knowledge extraction}.
\newblock In \emph{Forty-first International Conference on Machine Learning}.

\bibitem[{Gou et~al.(2024)Gou, Shao, Gong, yelong shen, Yang, Duan, and
  Chen}]{CRITIC}
Zhibin Gou, Zhihong Shao, Yeyun Gong, yelong shen, Yujiu Yang, Nan Duan, and
  Weizhu Chen. 2024.
\newblock \href {https://openreview.net/forum?id=Sx038qxjek} {{CRITIC}: Large
  language models can self-correct with tool-interactive critiquing}.
\newblock In \emph{The Twelfth International Conference on Learning
  Representations}.

\bibitem[{Honovich et~al.(2021)Honovich, Choshen, Aharoni, Neeman, Szpektor,
  and Abend}]{Q2}
Or~Honovich, Leshem Choshen, Roee Aharoni, Ella Neeman, Idan Szpektor, and Omri
  Abend. 2021.
\newblock \href {https://doi.org/10.18653/v1/2021.emnlp-main.619} {$q^{2}$:
  {E}valuating factual consistency in knowledge-grounded dialogues via question
  generation and question answering}.
\newblock In \emph{Proceedings of the 2021 Conference on Empirical Methods in
  Natural Language Processing}, Online and Punta Cana, Dominican Republic.
  Association for Computational Linguistics.

\bibitem[{Huang et~al.(2023)Huang, Gu, Hou, Wu, Wang, Yu, and
  Han}]{self-improve}
Jiaxin Huang, Shixiang Gu, Le~Hou, Yuexin Wu, Xuezhi Wang, Hongkun Yu, and
  Jiawei Han. 2023.
\newblock \href {https://doi.org/10.18653/v1/2023.emnlp-main.67} {Large
  language models can self-improve}.
\newblock In \emph{Proceedings of the 2023 Conference on Empirical Methods in
  Natural Language Processing}, pages 1051--1068, Singapore. Association for
  Computational Linguistics.

\bibitem[{Jiang et~al.(2024)Jiang, Li, Zhang, Huang, Lin, and
  Chen}]{TIGERScore}
Dongfu Jiang, Yishan Li, Ge~Zhang, Wenhao Huang, Bill~Yuchen Lin, and Wenhu
  Chen. 2024.
\newblock \href {https://arxiv.org/abs/2310.00752} {Tigerscore: Towards
  building explainable metric for all text generation tasks}.
\newblock \emph{Preprint}, arXiv:2310.00752.

\bibitem[{Kang et~al.(2024{\natexlab{a}})Kang, Ni, and Yao}]{EVER}
Haoqiang Kang, Juntong Ni, and Huaxiu Yao. 2024{\natexlab{a}}.
\newblock \href {https://arxiv.org/abs/2311.09114} {Ever: Mitigating
  hallucination in large language models through real-time verification and
  rectification}.
\newblock \emph{Preprint}, arXiv:2311.09114.

\bibitem[{Kang et~al.(2024{\natexlab{b}})Kang, Wallace, Tomlin, Kumar, and
  Levine}]{unfamiliar-finetuning}
Katie Kang, Eric Wallace, Claire Tomlin, Aviral Kumar, and Sergey Levine.
  2024{\natexlab{b}}.
\newblock \href {https://arxiv.org/abs/2403.05612} {Unfamiliar finetuning
  examples control how language models hallucinate}.
\newblock \emph{Preprint}, arXiv:2403.05612.

\bibitem[{Kim et~al.(2024)Kim, Shin, Cho, Jang, Longpre, Lee, Yun, Shin, Kim,
  Thorne, and Seo}]{prometheus}
Seungone Kim, Jamin Shin, Yejin Cho, Joel Jang, Shayne Longpre, Hwaran Lee,
  Sangdoo Yun, Seongjin Shin, Sungdong Kim, James Thorne, and Minjoon Seo.
  2024.
\newblock \href {https://openreview.net/forum?id=8euJaTveKw} {Prometheus:
  Inducing fine-grained evaluation capability in language models}.
\newblock In \emph{The Twelfth International Conference on Learning
  Representations}.

\bibitem[{Lebret et~al.(2016)Lebret, Grangier, and Auli}]{wikibio}
R{\'e}mi Lebret, David Grangier, and Michael Auli. 2016.
\newblock Neural text generation from structured data with application to the
  biography domain.
\newblock In \emph{Proceedings of the 2016 Conference on Empirical Methods in
  Natural Language Processing}. Association for Computational Linguistics.

\bibitem[{Lewis et~al.(2020)Lewis, Perez, Piktus, Petroni, Karpukhin, Goyal,
  K\"{u}ttler, Lewis, Yih, Rockt\"{a}schel, Riedel, and Kiela}]{RAG}
Patrick Lewis, Ethan Perez, Aleksandra Piktus, Fabio Petroni, Vladimir
  Karpukhin, Naman Goyal, Heinrich K\"{u}ttler, Mike Lewis, Wen-tau Yih, Tim
  Rockt\"{a}schel, Sebastian Riedel, and Douwe Kiela. 2020.
\newblock Retrieval-augmented generation for knowledge-intensive nlp tasks.
\newblock In \emph{Proceedings of the 34th International Conference on Neural
  Information Processing Systems}, NIPS '20.

\bibitem[{Li et~al.(2024{\natexlab{a}})Li, Sun, Yuan, Fan, hai zhao, and
  Liu}]{AutoJ}
Junlong Li, Shichao Sun, Weizhe Yuan, Run-Ze Fan, hai zhao, and Pengfei Liu.
  2024{\natexlab{a}}.
\newblock \href {https://openreview.net/forum?id=gtkFw6sZGS} {Generative judge
  for evaluating alignment}.
\newblock In \emph{The Twelfth International Conference on Learning
  Representations}.

\bibitem[{Li et~al.(2024{\natexlab{b}})Li, Chen, Ren, Cheng, Zhao, Nie, and
  Wen}]{empirical-factuality}
Junyi Li, Jie Chen, Ruiyang Ren, Xiaoxue Cheng, Xin Zhao, Jian-Yun Nie, and
  Ji-Rong Wen. 2024{\natexlab{b}}.
\newblock \href {https://aclanthology.org/2024.acl-long.586} {The dawn after
  the dark: An empirical study on factuality hallucination in large language
  models}.
\newblock In \emph{Proceedings of the 62nd Annual Meeting of the Association
  for Computational Linguistics (Volume 1: Long Papers)}, pages 10879--10899,
  Bangkok, Thailand. Association for Computational Linguistics.

\bibitem[{Li et~al.(2023{\natexlab{a}})Li, Cheng, Zhao, Nie, and
  Wen}]{halueval}
Junyi Li, Xiaoxue Cheng, Xin Zhao, Jian-Yun Nie, and Ji-Rong Wen.
  2023{\natexlab{a}}.
\newblock {H}alu{E}val: A large-scale hallucination evaluation benchmark for
  large language models.
\newblock In \emph{Proceedings of the 2023 Conference on Empirical Methods in
  Natural Language Processing}. Association for Computational Linguistics.

\bibitem[{Li et~al.(2024{\natexlab{c}})Li, Patel, Vi\'{e}gas, Pfister, and
  Wattenberg}]{ITI}
Kenneth Li, Oam Patel, Fernanda Vi\'{e}gas, Hanspeter Pfister, and Martin
  Wattenberg. 2024{\natexlab{c}}.
\newblock Inference-time intervention: eliciting truthful answers from a
  language model.
\newblock In \emph{Proceedings of the 37th International Conference on Neural
  Information Processing Systems}, NIPS '23, Red Hook, NY, USA.

\bibitem[{Li et~al.(2023{\natexlab{b}})Li, Zhang, Dubois, Taori, Gulrajani,
  Guestrin, Liang, and Hashimoto}]{alpaca_eval}
Xuechen Li, Tianyi Zhang, Yann Dubois, Rohan Taori, Ishaan Gulrajani, Carlos
  Guestrin, Percy Liang, and Tatsunori~B. Hashimoto. 2023{\natexlab{b}}.
\newblock Alpacaeval: An automatic evaluator of instruction-following models.
\newblock \url{https://github.com/tatsu-lab/alpaca_eval}.

\bibitem[{Lin et~al.(2022)Lin, Hilton, and Evans}]{truthfulqa}
Stephanie Lin, Jacob Hilton, and Owain Evans. 2022.
\newblock \href {https://doi.org/10.18653/v1/2022.acl-long.229}
  {{T}ruthful{QA}: Measuring how models mimic human falsehoods}.
\newblock In \emph{Proceedings of the 60th Annual Meeting of the Association
  for Computational Linguistics (Volume 1: Long Papers)}, pages 3214--3252,
  Dublin, Ireland. Association for Computational Linguistics.

\bibitem[{Madaan et~al.(2023)Madaan, Tandon, Gupta, Hallinan, Gao, Wiegreffe,
  Alon, Dziri, Prabhumoye, Yang, Gupta, Majumder, Hermann, Welleck,
  Yazdanbakhsh, and Clark}]{self-refine}
Aman Madaan, Niket Tandon, Prakhar Gupta, Skyler Hallinan, Luyu Gao, Sarah
  Wiegreffe, Uri Alon, Nouha Dziri, Shrimai Prabhumoye, Yiming Yang, Shashank
  Gupta, Bodhisattwa~Prasad Majumder, Katherine Hermann, Sean Welleck, Amir
  Yazdanbakhsh, and Peter Clark. 2023.
\newblock \href {https://openreview.net/forum?id=S37hOerQLB} {Self-refine:
  Iterative refinement with self-feedback}.
\newblock In \emph{Thirty-seventh Conference on Neural Information Processing
  Systems}.

\bibitem[{Maynez et~al.(2020)Maynez, Narayan, Bohnet, and
  Mcdonald}]{xsum-hallucination}
Joshua Maynez, Shashi Narayan, Bernd Bohnet, and Ryan~Thomas Mcdonald. 2020.
\newblock On faithfulness and factuality in abstractive summarization.
\newblock In \emph{Proceedings of the 58th Annual Meeting of the Association
  for Computational Linguistics}, pages 1906--1919, Online.

\bibitem[{Menick et~al.(2022)Menick, Trebacz, Mikulik, Aslanides, Song,
  Chadwick, Glaese, Young, Campbell-Gillingham, Irving, and McAleese}]{RLHF-QA}
Jacob Menick, Maja Trebacz, Vladimir Mikulik, John Aslanides, Francis Song,
  Martin Chadwick, Mia Glaese, Susannah Young, Lucy Campbell-Gillingham,
  Geoffrey Irving, and Nat McAleese. 2022.
\newblock \href {https://arxiv.org/abs/2203.11147} {Teaching language models to
  support answers with verified quotes}.
\newblock \emph{Preprint}, arXiv:2203.11147.

\bibitem[{Min et~al.(2023)Min, Krishna, Lyu, Lewis, Yih, Koh, Iyyer,
  Zettlemoyer, and Hajishirzi}]{factscore}
Sewon Min, Kalpesh Krishna, Xinxi Lyu, Mike Lewis, Wen-tau Yih, Pang Koh, Mohit
  Iyyer, Luke Zettlemoyer, and Hannaneh Hajishirzi. 2023.
\newblock \href {https://doi.org/10.18653/v1/2023.emnlp-main.741}
  {{FA}ct{S}core: Fine-grained atomic evaluation of factual precision in long
  form text generation}.
\newblock In \emph{Proceedings of the 2023 Conference on Empirical Methods in
  Natural Language Processing}, pages 12076--12100, Singapore. Association for
  Computational Linguistics.

\bibitem[{Mishra et~al.(2024)Mishra, Asai, Balachandran, Wang, Neubig,
  Tsvetkov, and Hajishirzi}]{FAVA}
Abhika Mishra, Akari Asai, Vidhisha Balachandran, Yizhong Wang, Graham Neubig,
  Yulia Tsvetkov, and Hannaneh Hajishirzi. 2024.
\newblock \href {https://openreview.net/forum?id=dJMTn3QOWO} {Fine-grained
  hallucination detection and editing for language models}.
\newblock In \emph{First Conference on Language Modeling}.

\bibitem[{Narayan et~al.(2018)Narayan, Cohen, and Lapata}]{xsum}
Shashi Narayan, Shay~B. Cohen, and Mirella Lapata. 2018.
\newblock \href {https://aclanthology.org/D18-1206} {Don{'}t give me the
  details, just the summary! topic-aware convolutional neural networks for
  extreme summarization}.
\newblock In \emph{Proceedings of the 2018 Conference on Empirical Methods in
  Natural Language Processing}. Association for Computational Linguistics.

\bibitem[{Ni et~al.(2022)Ni, Qu, Lu, Dai, Hernandez~Abrego, Ma, Zhao, Luan,
  Hall, Chang, and Yang}]{gtr}
Jianmo Ni, Chen Qu, Jing Lu, Zhuyun Dai, Gustavo Hernandez~Abrego, Ji~Ma,
  Vincent Zhao, Yi~Luan, Keith Hall, Ming-Wei Chang, and Yinfei Yang. 2022.
\newblock \href {https://aclanthology.org/2022.emnlp-main.669} {Large dual
  encoders are generalizable retrievers}.
\newblock In \emph{Proceedings of the 2022 Conference on Empirical Methods in
  Natural Language Processing}. Association for Computational Linguistics.

\bibitem[{Niu et~al.(2024)Niu, Wu, Zhu, Xu, Shum, Zhong, Song, and
  Zhang}]{ragtruth}
Cheng Niu, Yuanhao Wu, Juno Zhu, Siliang Xu, KaShun Shum, Randy Zhong, Juntong
  Song, and Tong Zhang. 2024.
\newblock {RAGT}ruth: A hallucination corpus for developing trustworthy
  retrieval-augmented language models.
\newblock In \emph{Proceedings of the 62nd Annual Meeting of the Association
  for Computational Linguistics (Volume 1: Long Papers)}. Association for
  Computational Linguistics.

\bibitem[{Ouyang et~al.(2024)Ouyang, Wu, Jiang, Almeida, Wainwright, Mishkin,
  Zhang, Agarwal, Slama, Ray, Schulman, Hilton, Kelton, Miller, Simens, Askell,
  Welinder, Christiano, Leike, and Lowe}]{InstructGPT}
Long Ouyang, Jeff Wu, Xu~Jiang, Diogo Almeida, Carroll~L. Wainwright, Pamela
  Mishkin, Chong Zhang, Sandhini Agarwal, Katarina Slama, Alex Ray, John
  Schulman, Jacob Hilton, Fraser Kelton, Luke Miller, Maddie Simens, Amanda
  Askell, Peter Welinder, Paul Christiano, Jan Leike, and Ryan Lowe. 2024.
\newblock Training language models to follow instructions with human feedback.
\newblock In \emph{Proceedings of the 36th International Conference on Neural
  Information Processing Systems}, NIPS '22. Curran Associates Inc.

\bibitem[{Pagnoni et~al.(2021)Pagnoni, Balachandran, and Tsvetkov}]{FRANK}
Artidoro Pagnoni, Vidhisha Balachandran, and Yulia Tsvetkov. 2021.
\newblock \href {https://doi.org/10.18653/v1/2021.naacl-main.383}
  {Understanding factuality in abstractive summarization with {FRANK}: A
  benchmark for factuality metrics}.
\newblock In \emph{Proceedings of the 2021 Conference of the North American
  Chapter of the Association for Computational Linguistics: Human Language
  Technologies}, pages 4812--4829, Online. Association for Computational
  Linguistics.

\bibitem[{Rafailov et~al.(2023)Rafailov, Sharma, Mitchell, Ermon, Manning, and
  Finn}]{DPO}
Rafael Rafailov, Archit Sharma, Eric Mitchell, Stefano Ermon, Christopher~D.
  Manning, and Chelsea Finn. 2023.
\newblock \href {https://arxiv.org/abs/2305.18290} {Direct preference
  optimization: Your language model is secretly a reward model}.
\newblock \emph{Preprint}, arXiv:2305.18290.

\bibitem[{Saunders et~al.(2022)Saunders, Yeh, Wu, Bills, Ouyang, Ward, and
  Leike}]{self-critiquing}
William Saunders, Catherine Yeh, Jeff Wu, Steven Bills, Long Ouyang, Jonathan
  Ward, and Jan Leike. 2022.
\newblock \href {https://arxiv.org/abs/2206.05802} {Self-critiquing models for
  assisting human evaluators}.
\newblock \emph{Preprint}, arXiv:2206.05802.

\bibitem[{Schuster et~al.(2021)Schuster, Fisch, and Barzilay}]{vitaminc}
Tal Schuster, Adam Fisch, and Regina Barzilay. 2021.
\newblock \href {https://aclanthology.org/2021.naacl-main.52} {Get your vitamin
  {C}! robust fact verification with contrastive evidence}.
\newblock In \emph{Proceedings of the 2021 Conference of the North American
  Chapter of the Association for Computational Linguistics: Human Language
  Technologies}. Association for Computational Linguistics.

\bibitem[{See et~al.(2017)See, Liu, and Manning}]{cnndm}
Abigail See, Peter~J. Liu, and Christopher~D. Manning. 2017.
\newblock \href {https://www.aclweb.org/anthology/P17-1099} {Get to the point:
  Summarization with pointer-generator networks}.
\newblock In \emph{Proceedings of the 55th Annual Meeting of the Association
  for Computational Linguistics (Volume 1: Long Papers)}. Association for
  Computational Linguistics.

\bibitem[{Shi et~al.(2024)Shi, Han, Lewis, Tsvetkov, Zettlemoyer, and
  Yih}]{CAD}
Weijia Shi, Xiaochuang Han, Mike Lewis, Yulia Tsvetkov, Luke Zettlemoyer, and
  Wen-tau Yih. 2024.
\newblock \href {https://aclanthology.org/2024.naacl-short.69} {Trusting your
  evidence: Hallucinate less with context-aware decoding}.
\newblock In \emph{Proceedings of the 2024 Conference of the North American
  Chapter of the Association for Computational Linguistics: Human Language
  Technologies (Volume 2: Short Papers)}, pages 783--791, Mexico City, Mexico.
  Association for Computational Linguistics.

\bibitem[{Tang et~al.(2024)Tang, Laban, and Durrett}]{llm-aggrefact}
Liyan Tang, Philippe Laban, and Greg Durrett. 2024.
\newblock \href {https://arxiv.org/abs/2404.10774} {Minicheck: Efficient
  fact-checking of llms on grounding documents}.
\newblock \emph{Preprint}, arXiv:2404.10774.

\bibitem[{Tian et~al.(2024)Tian, Mitchell, Yao, Manning, and Finn}]{FactTune}
Katherine Tian, Eric Mitchell, Huaxiu Yao, Christopher~D Manning, and Chelsea
  Finn. 2024.
\newblock \href {https://openreview.net/forum?id=WPZ2yPag4K} {Fine-tuning
  language models for factuality}.
\newblock In \emph{The Twelfth International Conference on Learning
  Representations}.

\bibitem[{Vu et~al.(2024)Vu, Krishna, Alzubi, Tar, Faruqui, and Sung}]{FLAMe}
Tu~Vu, Kalpesh Krishna, Salaheddin Alzubi, Chris Tar, Manaal Faruqui, and
  Yun-Hsuan Sung. 2024.
\newblock \href {https://arxiv.org/abs/2407.10817} {Foundational autoraters:
  Taming large language models for better automatic evaluation}.
\newblock \emph{Preprint}, arXiv:2407.10817.

\bibitem[{Wang et~al.(2020)Wang, Cho, and Lewis}]{QAGS}
Alex Wang, Kyunghyun Cho, and Mike Lewis. 2020.
\newblock \href {https://doi.org/10.18653/v1/2020.acl-main.450} {Asking and
  answering questions to evaluate the factual consistency of summaries}.
\newblock In \emph{Proceedings of the 58th Annual Meeting of the Association
  for Computational Linguistics}, pages 5008--5020, Online. Association for
  Computational Linguistics.

\bibitem[{Wang et~al.(2023)Wang, Yu, Tan, O'Brien, Pasunuru, Dwivedi-Yu,
  Golovneva, Zettlemoyer, Fazel-Zarandi, and Celikyilmaz}]{shepherd}
Tianlu Wang, Ping Yu, Xiaoqing~Ellen Tan, Sean O'Brien, Ramakanth Pasunuru,
  Jane Dwivedi-Yu, Olga Golovneva, Luke Zettlemoyer, Maryam Fazel-Zarandi, and
  Asli Celikyilmaz. 2023.
\newblock \href {https://arxiv.org/abs/2308.04592} {Shepherd: A critic for
  language model generation}.
\newblock \emph{Preprint}, arXiv:2308.04592.

\bibitem[{Wei et~al.(2024)Wei, Yang, Song, Lu, Hu, Huang, Tran, Peng, Liu,
  Huang, Du, and Le}]{longform}
Jerry Wei, Chengrun Yang, Xinying Song, Yifeng Lu, Nathan Hu, Jie Huang, Dustin
  Tran, Daiyi Peng, Ruibo Liu, Da~Huang, Cosmo Du, and Quoc~V. Le. 2024.
\newblock \href {https://arxiv.org/abs/2403.18802} {Long-form factuality in
  large language models}.
\newblock \emph{Preprint}, arXiv:2403.18802.

\bibitem[{Welleck et~al.(2023)Welleck, Lu, West, Brahman, Shen, Khashabi, and
  Choi}]{self-correct}
Sean Welleck, Ximing Lu, Peter West, Faeze Brahman, Tianxiao Shen, Daniel
  Khashabi, and Yejin Choi. 2023.
\newblock \href {https://openreview.net/forum?id=hH36JeQZDaO} {Generating
  sequences by learning to self-correct}.
\newblock In \emph{The Eleventh International Conference on Learning
  Representations}.

\bibitem[{Xie et~al.(2024)Xie, Zhang, Cheng, Liu, Gero, Wong, Naumann, Poon,
  and Rose}]{doclens}
Yiqing Xie, Sheng Zhang, Hao Cheng, Pengfei Liu, Zelalem Gero, Cliff Wong,
  Tristan Naumann, Hoifung Poon, and Carolyn Rose. 2024.
\newblock Doclens: Multi-aspect fine-grained evaluation for medical text
  generation.
\newblock In \emph{Proceedings of the 62nd Annual Meeting of the Association
  for Computational Linguistics}.

\bibitem[{Xu et~al.(2024)Xu, Zhu, Zhao, Pan, Li, and Wang}]{self-bias}
Wenda Xu, Guanglei Zhu, Xuandong Zhao, Liangming Pan, Lei Li, and William~Yang
  Wang. 2024.
\newblock \href {https://arxiv.org/abs/2402.11436} {Pride and prejudice: Llm
  amplifies self-bias in self-refinement}.
\newblock \emph{Preprint}, arXiv:2402.11436.

\bibitem[{Zhang et~al.(2024)Zhang, Peng, Tian, Zhou, Jin, Song, Mi, and
  Meng}]{self-eval-skt}
Xiaoying Zhang, Baolin Peng, Ye~Tian, Jingyan Zhou, Lifeng Jin, Linfeng Song,
  Haitao Mi, and Helen Meng. 2024.
\newblock \href {https://doi.org/10.18653/v1/2024.acl-long.107} {Self-alignment
  for factuality: Mitigating hallucinations in {LLM}s via self-evaluation}.
\newblock In \emph{Proceedings of the 62nd Annual Meeting of the Association
  for Computational Linguistics (Volume 1: Long Papers)}, Bangkok, Thailand.
  Association for Computational Linguistics.

\bibitem[{Zhang et~al.(2023)Zhang, Li, Cui, Cai, Liu, Fu, Huang, Zhao, Zhang,
  Chen, Wang, Luu, Bi, Shi, and Shi}]{hallucination-survey}
Yue Zhang, Yafu Li, Leyang Cui, Deng Cai, Lemao Liu, Tingchen Fu, Xinting
  Huang, Enbo Zhao, Yu~Zhang, Yulong Chen, Longyue Wang, Anh~Tuan Luu, Wei Bi,
  Freda Shi, and Shuming Shi. 2023.
\newblock \href {https://arxiv.org/abs/2309.01219} {Siren's song in the ai
  ocean: A survey on hallucination in large language models}.
\newblock \emph{Preprint}, arXiv:2309.01219.

\bibitem[{Zhou et~al.(2018)Zhou, Prabhumoye, and Black}]{cmu-dog}
Kangyan Zhou, Shrimai Prabhumoye, and Alan~W Black. 2018.
\newblock \href {https://arxiv.org/abs/1809.07358} {A dataset for document
  grounded conversations}.
\newblock \emph{Preprint}, arXiv:1809.07358.

\bibitem[{Ziegler et~al.(2020)Ziegler, Stiennon, Wu, Brown, Radford, Amodei,
  Christiano, and Irving}]{RLHF}
Daniel~M. Ziegler, Nisan Stiennon, Jeffrey Wu, Tom~B. Brown, Alec Radford,
  Dario Amodei, Paul Christiano, and Geoffrey Irving. 2020.
\newblock \href {https://arxiv.org/abs/1909.08593} {Fine-tuning language models
  from human preferences}.
\newblock \emph{Preprint}, arXiv:1909.08593.

\end{thebibliography}

\clearpage
\appendix

\section{Implementation Details}
\label{sec:implementation_details}
\start{Evaluator Training: Obtaining Critiques}
As introduced in \S\ref{sec:evaluator_training}, for each prompt in the training set, we prompt Llama3-70B-chat to generate 10 candidate judgments, each containing both the critique and the label. If one of the 10 candidates has the same label as the ground truth label, we add the critique generated by this candidate to the training set. Otherwise, we discard the whole example.

Most existing datasets only have binary labels (``factual'' or ``non-factual'') and our label space includes three label classes. 
We hence match ``supported'' with ``factual'' and match both ``contradictory'' and ``unverified'' to ``non-factual''.

\start{Evaluator Training: Obtaining Source Documents by Tools}
We also augment the source documents by calling three tools: Bing Web Search API, an offline copy of Wikipedia, and Google Knowledge Graph API, which represents three types of tools: search engine, knowledge base, and knowledge graph.
We first put the documents obtained by all the tools in a document list, then we rerank the documents based on the cosine similarity between their text embeddings and the claim's embedding, where we use GTR-T5-Large~\citep{gtr} as the encoder.
We provide the evaluator with the top-5 documents.

To call Bing Web Search API, we first prompt Llama3-70B-chat to generate a search query with the instruction: ``You are given a STATEMENT. Your task is to write one SEARCH QUERY to find evidence supporting or disproving the STATEMENT.''
Then we call the Bing Search API, which returns 5 search results for each query. Each result contains the URL and a short snippet.
We further scrape each URL to obtain the full content of the webpage and chunk the content (with 512 as the chunk size).
Finally, we add all the chunks and snippets to the document list. 

To call Wikipedia, we download an offline copy of Wikipedia (the 2023/04/01 version). Similar to search query generation, we prompt Llama3-70B-chat to generate a list of possible Wikipage names. For each generated name, we retrieve the top-3 Wikipage based on cosine similarity of the pagename embeddings. We chunk the content of all the retrieved Wikipages and add the chunks to the document list.

To call the Google Knowledge Graph API, we prompt Llama3-70B-chat to generate a list of entities for each claim and add the top-1 returned result to the document list (if any).

\start{Generator Training: Baselines and Ablations}
For all the baselines and ablations, we use Llama3-8B-chat as initialization, set $N=5$, use Llama3-70B-chat to decompose the responses into facts, and call the same tools for evaluation (Bing Search, Wikipedia, Google Knowledge Graph).
We use Llama3-8B-chat as the evaluator for both baseline methods.
For our method and EVER, we revise at most 3 false claims (i.e., the ones judged as ``contradictory'' or ``unverified'') for each of the first three passages.

\begin{table*}[t]
\centering
\resizebox{0.85\linewidth}{!}{
\begin{tabular}{lcc}
\toprule
\bf Category & \bf Dataset Name & \bf Base Datasets \\ 
\midrule
\multirow{6}{*}{Summarization} & XSum Hallucination~\citep{xsum-hallucination} & XSum~\citep{xsum} \\
\cmidrule{2-3}
& \multirow{2}{*}{QAGS~\citep{QAGS}} & XSum~\citep{xsum} \\
& & CNN/DM~\citep{cnndm} \\
\cmidrule{2-3}
& \multirow{2}{*}{FRANK~\citep{FRANK}} & XSum~\citep{xsum} \\
& & CNN/DM~\citep{cnndm} \\
\cmidrule{2-3}
& RAGTruth~\citep{ragtruth} & CNN/DM~\citep{cnndm} \\

\midrule 
\multirow{2}{*}{Question Answering \quad} & RAGTruth~\citep{ragtruth} & MSMARCO~\citep{msmarco} \\
\cmidrule{2-3}
& FActSore~\citep{factscore} & WikiBio~\citep{wikibio} \\

\midrule 
\multirow{4}{*}{Dialogue} & Q-Square~\citep{Q2} & Wizard of Wikipedia~\citep{wow} \\
\cmidrule{2-3}
& FaithDial~\citep{faithdial} & Wizard of Wikipedia~\citep{wow} \\
\cmidrule{2-3}
& \multirow{2}{*}{BEGIN~\citep{begin}} & Wizard of Wikipedia~\citep{wow} \\
& & CMU-DoG~\citep{cmu-dog} \\

\bottomrule
\end{tabular}
}
\caption{
The list of training datasets we used to train \modelname.
}
\label{tab:evaluator_train_data}
\end{table*}

\section{Dataset Details}
\label{sec:data_details}
\start{Evaluator Training Data}
We provide the list of datasets we used to train \modelname in \autoref{tab:evaluator_train_data}. 
All the datasets are open-sourced on HugginFace or GitHub.
We only focus on datasets with \textit{human judgments} on \textit{model-generated responses} or claims decomposed from model responses, excluding synthetic datasets such as HaluEval~\citep{halueval}, sentence revision-based datasets such as VitaminC~\citep{vitaminc}, etc.

\start{Generator Training and Testing Data}
We only require a prompt set to finetune the generator. Following previous work~\citep{FactTune,EVER}, we conduct experiments on the FActScore dataset~\citep{factscore}.

Unlike \citet{EVER} that uses the same set of prompts for training and testing, to evaluate the generator's generalizability to unseen prompts, we 
follow \citet{FactTune} and use different prompts for training and testing.
Since \citet{FactTune} does not release their train-test split, we make our own split by randomly dividing the unlabeled subset of FActScore into 400 training and 100 testing prompts\footnote{We release our split of training testing prompts at:
\url{https://drive.google.com/drive/folders/1GsTmoh1t1jInSrUcgej1kZWG7KNXDFL4?usp=sharing}.}.

\section{Case Studies}
\label{sec:case_studies}

\start{Evaluator Training: Data Augmentation}
\autoref{tab:case_1} shows an example of our generated textual critique, which is aligned with the ground truth label. 
\autoref{tab:case_2} and \autoref{tab:case_3} show two examples where we obtain additional source documents by calling tools. We can see that compared to the existing document, the tool-extracted documents are more diverse in terms of sources, content, and formats, and are still correctly labeled.

\begin{table*}[!t]
\centering

\resizebox{\textwidth}{!}{
\begin{tabular}{p{20.1cm}}
\toprule
\dlrow \textit{\textbf{Source Document} (Dataset: FRANK-xsum)}  \\
Title: News Articles \\
Text: They believe ministers are placing too much emphasis on the environment at the expense of trees grown for timber.
Britain is currently the world's third largest importer of wood.
Ministers said they were encouraging commercial forestry organisations to invest in woodland creation.
Conifer forests have been a familiar sight for half a century in Wales and have helped the timber industry grow.
But Confor, which promotes the forestry industry, warns that at least 16,000 hectares - or 40,000 acres - of commercial forest have been lost since 2001 and need to be re-planted to meet needs.
Half of the woodland is managed by Natural Resources Wales with the the other half by private companies.
As an industry, it is estimated to be worth more than Â£450m a year to the Welsh economy.
But there is a conflict with environmental policy.
\textbf{Successive ministers have wanted more native, broadleaf species of trees and better access for walkers and bike riders while conservationists claim commercial forests are bad for wildlife.}
However, some forests in Wales are difficult to reach and hard to harvest while transporting the timber through rural communities can bring companies into conflict with local residents.
The Welsh government wants a balance between the environmental, the economic and the social in how forestry is developed.
It points to its Glastir scheme which encourages farmers and other landowners to plant woodland.
It said its Woodlands for Wales forestry strategy "sets out how we want Wales to be known for its high quality woodlands that enhance the landscape, provide real social and community benefits, support thriving woodland-based industries and contribute to a better quality environment".
Martin Bishop, of Confor, which represents 200 businesses in Wales, believes the emphasis on environmental aspects is seen as "sexier" than commercial forestry.
He said the current system is also too bureaucratic for creating new woodland areas.
Mr Bishop said 90 hectares were planted last year but there needed to be 20,000 hectares a year to meet long-term targets set by ministers.
"There's a huge market for our timber. Every saw miller tells me he would double or treble production if the timber was there to feed that mill," he added. \\
\\
\midrule 

\dlrow \textit{\textbf{Claim}} \\
Conservationists are calling for a national project to protect forests in wales. \\
\\
\dlrow \textit{\textbf{Ground Truth Label}} \\
\colorR{Non-Factual} \\
\\
\midrule

\blrow \textit{\textbf{Our Generated Critique and Label}} \\
\textbf{Critique}: The document does not mention conservationists calling for a national project to protect forests in Wales. It only mentions that conservationists claim commercial forests are bad for wildlife, but it does not mention a specific call for a national project. \\
\\
\textbf{Label}: \colorR{Unverified} \\
\bottomrule
\end{tabular}
}
\caption{An example of the textual critique we generate. We highlight the most relevant sentence in the document. In this case, our generated critique aligns with the judgment label.
}
\label{tab:case_1}
\end{table*}

\begin{table*}[!t]
\centering

\resizebox{\textwidth}{!}{
\begin{tabular}{p{20.1cm}}
\toprule
\dlrow \textit{\textbf{Original Source Document} (Dataset: FRANK-CNN/DM)}  \\ 
Title: News Articles \\
Text: (CNN) Deputies rushed Kenneth Morgan Stancil III from court Thursday after the 20-year-old murder suspect swore at a judge and tried to flip over a table. Stancil is accused of killing an employee Monday at Wayne Community College in Goldsboro, North Carolina. Relatives have said victim Ron Lane was gay, CNN affiliate WNCN reported, and investigators are looking into whether the shooting was a hate crime. Authorities arrested Stancil after he was found sleeping on a Florida beach on Tuesday. Just a few minutes into Thursday's hearing on the first-degree murder charge he faces, Stancil snapped back at the judge after he was offered a court-appointed lawyer. \"No, I don't need one,\" said Stancil, who stood before the judge with his legs shackled and his arms handcuffed in front of him. \"You know what I'm saying? I knew I would get life anyway.\" Superior Court Judge Arnold O. Jones interjected, pointing out that the maximum sentence Stancil faces is the death penalty. \"Yes, I know that,\" Stancil fired back. \"But when I knew what I had to do and I knew when I got caught, you know, I knew in my mind that I could get life, I could get the death penalty. You know what I'm saying? Do you follow my topic? I would have killed you, you know what I'm saying, if you're a f---ing child molester.\" The judge told him not to swear. \"I don't give a f--- what you want,\" Stancil said, lunging forward and lifting up the table in front of him. Deputies quickly corralled him and hustled him from the courtroom. The hearing resumed about 25 minutes later, when Stancil was brought back into the courtroom, this time with his arms handcuffed behind him. When asked again by Jones whether he wanted a lawyer, his response was quick -- and calm. \"Yes, sir,\" he said. In an interview with CNN affiliate WRAL, Stancil described himself as a neo-Nazi and said he hates gay people \"with a passion.\" Stancil had worked for Lane, the school's print shop operator, as part of a work-study program, but was let go from the program in early March because of poor attendance, college officials said. \\
\\
\midrule 

\dlrow \textit{\textbf{Claim}} \\
Kenneth Morgan Stancil is accused of killing an employee at Wayne Community College in Goldsboro, North Carolina. \\
\\
\dlrow \textit{\textbf{Ground Truth Label}} \\
\colorG{Factual} \\
\\
\midrule

\blrow \textit{\textbf{Generated Queries to Call the Tools}} \\
\textbf{Search query for Bing Search API}: Kenneth Morgan Stancil Wayne Community College killing \\
\textbf{Queries for Wikipedia}: Kenneth Morgan Stancil; List of school shootings in the United States; Wayne Community College; \\
\textbf{Entities for Google Knowledge Graph API}: Kenneth Morgan Stancil \\
\\
\midrule

\blrow \textit{\textbf{Tool-extracted Source Documents} (after Reranking)} \\
Title: Wayne Community College shooter gets life sentence without parole \\
Text: GOLDSBORO, North Carolina (WTVD) -- \textbf{Kenneth Morgan Stancil III was sentenced Tuesday to life in prison without parole for the murder of 44-year-old Ron Lane on the campus of Wayne Community College in Goldsboro on April 13, 2015.}
Stancil entered the campus print shop on the third floor of the same building that houses the school library and cafeteria shortly after Lane arrived for work that day and shot him once with a pistol-grip 12-gauge shotgun.
...
\\
\\
Title: Stancil guilty in Wayne Community College murder trial - CBS17.com \\
Text: GOLDSBORO, N.C. (WNCN) – After deliberating for an hour and a half, \textbf{a Wayne County jury found Kenneth Morgan Stancil III guilty of first-degree murder.}
He will spend the rest of his life in prison.
During his closing argument Tuesday in a Wayne County court, the district attorney relied largely on Stancil’s own words, \textbf{painting him as confident and proud about his decision to kill a man at Wayne Community College.}
In his counterargument, Stancil’s attorney said the prosecution tried to scare the jury with Stancil’s possession of knives, guns, and white supremacist memorabilia, as well as his numerous tattoos. None of that proved he planned to commit the murder, the defense said.
\textbf{Neither side disputed that Stancil killed Ron Lane at Wayne Community College in 2015}. The prosecution contends it was first-degree murder, but the defense argues ...
\\
\\
Title: Man convicted in fatal Wayne Community College shooting - WRAL ... \\
Text: Man convicted in fatal Wayne Community College shooting
A white supremacist was sentenced Tuesday to life in prison without parole for gunning down a print shop operator at Wayne Community College two years ago.
\textbf{Jurors deliberated for about 90 minutes before convicting Kenneth Morgan Stancil III, 22, of first-degree murder in the April 13, 2015, death of Ron Lane in the school's library.}
Authorities said Stancil shot Lane with a 12-gauge pump shotgun before fleeing the area on a motorcycle. He was captured the next morning when police found him asleep on a beach in Volusia County, Fla.
...
\\
\\
Title: Wayne Community College killing suspect makes shocking allegations in ... \\
... \\
\midrule

\blrow \textit{\textbf{Our Generated Critique and Label}} \\
\textbf{Critique}: The statement is supported by the document. In all the documents, it is mentioned that Kenneth Morgan Stancil III is accused of killing Ron Lane, an employee at Wayne Community College in Goldsboro, North Carolina. \\
\\
\textbf{Label}: \colorG{Supported} \\
\bottomrule
\end{tabular}
}
\caption{An example where we obtain multiple relevant source documents by calling a multiplicity of tools.
}
\label{tab:case_2}
\end{table*}

\begin{table*}[!t]
\centering

\resizebox{\textwidth}{!}{
\begin{tabular}{p{20.1cm}}
\toprule
\dlrow \textit{\textbf{Original Source Document} (Dataset: FActScore)}  \\ 
Title: Chadwick Boseman \\
Text: Chadwick Boseman Chadwick Aaron Boseman \textbf{(; November 29, 1976August 28, 2020)} was an American actor. During his two-decade career, Boseman received multiple accolades, including two Screen Actors Guild Awards, a Golden Globe Award, a Critics' Choice Movie Award, a Primetime Emmy Award, and an Academy Award nomination. After studying directing at Howard University, Boseman began his career in theatre, winning a Drama League Directing Fellowship and an acting AUDELCO ...
\\
\\
\midrule 

\dlrow \textit{\textbf{Claim}} \\
Chadwick Boseman was born on November 29, 1977. \\
\\
\dlrow \textit{\textbf{Ground Truth Label}} \\
\colorR{Non-Factual} \\
\\
\midrule

\blrow \textit{\textbf{Generated Queries to Call the Tools}} \\
\textbf{Search query for Bing Search API}: Chadwick Boseman date of birth \\
\textbf{Queries for Wikipedia}: Chadwick Boseman \\
\textbf{Entities for Google Knowledge Graph API}: Chadwick Boseman Stancil \\
\\
\midrule

\blrow \textit{\textbf{Tool-extracted Source Documents} (after Reranking)} \\
Title: knowledge graph result \\
Text: Chadwick Boseman: American actor. Chadwick Aaron Boseman was an American actor, known for portraying African-American historical figures and the fictional African superhero Black Panther. \textbf{Chadwick Boseman Born: November 29, 1976, Anderson, SC.} Chadwick Boseman Died: August 28, 2020 (age 43\u00a0years), Los Angeles, CA. Chadwick Boseman Spouse: Taylor Simone Ledward. Chadwick Boseman Height: 6\u2032 0\u2033. Chadwick Boseman Parents: Leroy Boseman and Carolyn Boseman. Chadwick Boseman Alma mater: Howard University (BFA).
...
\\
\\
Title: Chadwick Boseman | Biography, Movies, Black Panther, \& Death
text: Chadwick Boseman \\
Text: Our editors will review what you’ve submitted and determine whether to revise the article. \\
- Chadwick Aaron Boseman \\
\textbf{- Born: November 29, 1976, Anderson, South Carolina, U.S.} \\
- Died: August 28, 2020, Los Angeles, California (aged 43) \\
- Notable Family Members: son of Leroy Boseman ... \\
Chadwick Boseman (born November 29, 1976, Anderson, South Carolina, U.S.—died August 28, 2020, Los Angeles, California) was an American actor and playwright who became a highly respected movie star with several iconic roles, notably that of T’Challa/Black Panther in the groundbreaking film Black Panther (2018).
...
\\
\\
Title: Chadwick Boseman - Wikipedia \\
Text: Chadwick Boseman Chadwick Aaron Boseman \textbf{(; November 29, 1976August 28, 2020)} was an American actor. During his two-decade career, Boseman received multiple accolades, including two Screen Actors Guild Awards, a Golden Globe Award, a Critics' Choice Movie Award, a Primetime Emmy Award, and an Academy Award nomination. After studying directing at Howard University, Boseman began his career in theatre, winning a Drama League Directing Fellowship and an acting AUDELCO ...
\\
\\
Title: Chadwick Boseman | Biography, Movies, Black Panther, \& Death ... \\
... \\
\midrule

\blrow \textit{\textbf{Our Generated Critique and Label}} \\
\textbf{Critique}: The statement claims that Chadwick Boseman was born on November 29, 1977. However, all four documents consistently state that Chadwick Boseman was born on November 29, 1976, not 1977. \\
\\
\textbf{Label}: \colorR{Contradictory} \\
\bottomrule
\end{tabular}
}
\caption{An example where we obtain multiple relevant source documents by calling a multiplicity of tools. The tool-extracted documents all contain the information about Chadwick's birthday, but have diverse formats.
}
\label{tab:case_3}
\end{table*}

\end{document}